\documentclass[sigconf]{acmart}
\AtBeginDocument{%
  }

\copyrightyear{2026}
\acmYear{2026}
\setcopyright{cc}
\setcctype{by}
\acmConference[MM '26]{Proceedings of the 34th ACM International Conference on Multimedia}{November 10--14, 2026}{Rio de Janeiro, Brazil}
\acmBooktitle{Proceedings of the 34th ACM International Conference on Multimedia (MM '26), November 10--14, 2026, Rio de Janeiro, Brazil}
\acmDOI{10.1145/3767308.3835210}
\acmISBN{979-8-4007-2213-4/2026/11}

\acmSubmissionID{1783}

\usepackage{xcolor}
\usepackage{colortbl}
\usepackage{booktabs}   %
\usepackage{multirow}   %
\usepackage{array}      %
\usepackage{mathtools}  %
\usepackage{graphicx}
\usepackage{pifont}
\usepackage{bbding}

\newcommand{\equalcontrib}{\textsuperscript{\ensuremath{\dagger}}}
\newcommand{\corrauth}{\textsuperscript{\ensuremath{\ddagger}}}
\newcommand{\projlead}{\textsuperscript{\S}}
\definecolor{PositiveGreen}{RGB}{0, 145, 0}  %
\definecolor{NegativeRed}{RGB}{210, 0, 0}    %
\newcommand{\up}[1]{\ensuremath{\mathrlap{_{\textcolor{PositiveGreen}{+#1}}}}}
\newcommand{\down}[1]{\ensuremath{\mathrlap{_{\textcolor{NegativeRed}{-#1}}}}}

\begin{document}

\title{Latent-WAM: Latent World Action Modeling for End-to-End Autonomous Driving}

\author{Linbo Wang\equalcontrib}
\authornote{%
Work done during internship at Chongqing Chang’an Technology Co., Ltd. This work is supported by Beijing Natural Science
Foundation under Grant 4242052 and Beijing Nova Program (20240484562, 202604841268). \\
\equalcontrib Equal contribution.\hspace{0.8em}
\corrauth Corresponding author.\hspace{0.8em}
\projlead Project Leader.%
}
\affiliation{%
    \institution{MAIS-SKL, CASIA}
    \institution{Chongqing Chang’an Technology Co., Ltd.}
    \institution{Zhongguancun Academy}
    \city{Beijing}
    \country{China}
}
\email{wanglinbo2026@ia.ac.cn}

\author{Yupeng Zheng\equalcontrib}
\affiliation{%
    \institution{MAIS-SKL, CASIA}
    \city{Beijing}
    \country{China}
}
\email{zhengyupeng2022@ia.ac.cn}

\author{Qiang Chen}
\affiliation{%
    \institution{Chongqing Chang’an Technology Co., Ltd.}
    \city{Beijing}
    \country{China}
}
\email{chenqiang01@changan.com.cn}

\author{Shiwei Li}
\affiliation{%
    \institution{Chongqing Chang’an Technology Co., Ltd.}
    \city{Beijing}
    \country{China}
}
\email{lisw2@changan.com.cn}

\author{Yichen Zhang}
\affiliation{%
    \institution{MAIS-SKL, CASIA}
    \city{Beijing}
    \country{China}
}
\email{zhangyichen2026@ia.ac.cn}

\author{Zebin Xing}
\affiliation{%
    \institution{MAIS-SKL, CASIA}
    \city{Beijing}
    \country{China}
}
\email{xingzebin2024@ia.ac.cn}

\author{Qichao Zhang\corrauth}
\affiliation{%
    \institution{MAIS-SKL, CASIA}
    \city{Beijing}
    \country{China}
}
\email{zhangqichao2014@ia.ac.cn}

\author{Xiang Li}
\affiliation{%
    \institution{Tsinghua University}
    \city{Beijing}
    \country{China}
}
\email{l-x21@mails.tsinghua.edu.cn}

\author{Deheng Qian}
\affiliation{%
    \institution{Chongqing Chang’an Technology Co., Ltd.}
    \city{Beijing}
    \country{China}
}
\email{qiandeheng@changan.com.cn}

\author{Pengxuan Yang}
\affiliation{%
    \institution{MAIS-SKL, CASIA}
    \city{Beijing}
    \country{China}
}
\email{yangpengxuan2024@ia.ac.cn}

\author{Yihang Dong}
\affiliation{%
    \institution{Zhongguancun Academy}
    \city{Beijing}
    \country{China}
}
\email{2022190901025@std.uestc.edu.cn}

\author{Ce Hao\corrauth}
\affiliation{%
    \institution{Zhongguancun Academy}
    \city{Beijing}
    \country{China}
}
\email{cehao@berkeley.edu}

\author{Xiaoqing Ye\corrauth}
\affiliation{%
    \institution{Wuhan University}
    \city{Wuhan}
    \country{China}
}
\email{yxq@whu.edu.cn}

\author{Junyu Han}
\affiliation{%
    \institution{Chongqing Chang’an Technology Co., Ltd.}
    \city{Beijing}
    \country{China}
}
\email{hanjunyu@changan.com.cn}

\author{Yifeng Pan\projlead}
\affiliation{%
    \institution{Chongqing Chang’an Technology Co., Ltd.}
    \city{Beijing}
    \country{China}
}
\email{panyifeng@changan.com.cn}

\author{Dongbin Zhao\projlead}
\affiliation{%
    \institution{MAIS-SKL, CASIA}
    \institution{Zhongguancun Academy}
    \city{Beijing}
    \country{China}
}
\email{dongbin.zhao@ia.ac.cn}

\renewcommand{\shortauthors}{Linbo Wang et al.}

\begin{abstract}
We introduce Latent-WAM, an efficient end-to-end autonomous driving framework
that achieves strong trajectory planning through spatially-aware and
dynamics-informed latent world representations. Existing world-model-based
planners suffer from inadequately compressed representations, limited spatial
understanding, and underutilized temporal dynamics, resulting in sub-optimal
planning under constrained data and compute budgets. Latent-WAM addresses these
limitations with two core modules: a Spatial-Aware Compressive World Encoder
(SCWE) that distills geometric knowledge from a foundation model and compresses
multi-view images into compact scene tokens via learnable queries, and a Dynamic Latent World Model (DLWM) that employs a causal Transformer to
autoregressively predict future world status conditioned on historical visual
and ego status representations. Extensive experiments on NAVSIM v2 and HUGSIM
demonstrate new state-of-the-art results: 89.3 EPDMS on NAVSIM v2 and 28.9
HD-Score on HUGSIM, surpassing the best prior perception-free method by 3.2
EPDMS with significantly less training data and a compact 104M-parameter model. The code and model weights have been open-sourced at \textit{\url{https://github.com/yukki555/Latent-WAM}}.
\end{abstract}



\maketitle

\section{Introduction}
End-to-end autonomous driving has attracted considerable attention due to its data-driven nature and scalability.
Prior methods integrate perception and prediction into a unified differentiable network, extracting planning-relevant representations for end-to-end trajectory prediction~\cite{hu2023uniad, jiang2023vad}. However, these methods generally rely on complex auxiliary task designs and perception annotations, limiting further scaling of end-to-end driving algorithms. Large-scale studies~\cite{zheng2026datascaling} further show that end-to-end driving performance scales predictably with data, motivating more effective representation learning from limited data.

\begin{figure}
    \centering
    \includegraphics[width=\linewidth]{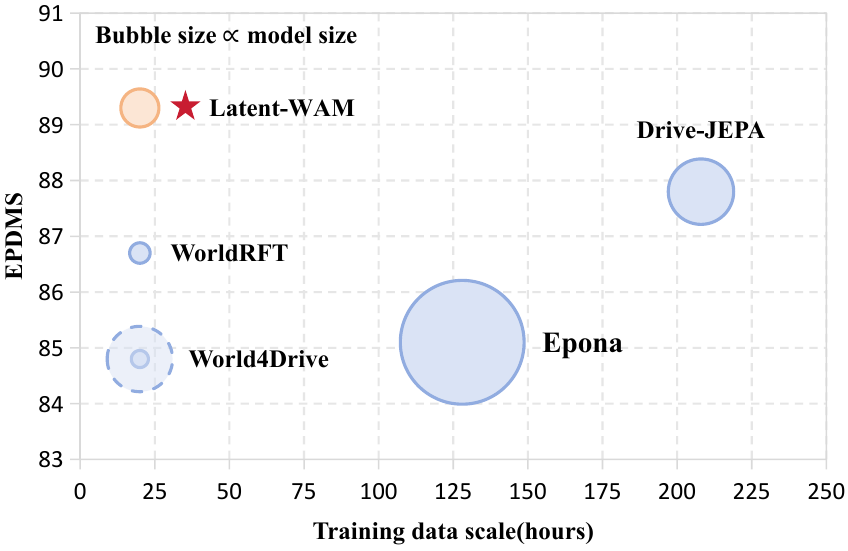}
    \caption{\textbf{Performance vs.\ training data scale on NAVSIM v2.} Bubble size indicates model parameters. Our
    Latent-WAM achieves the highest EPDMS with significantly fewer training data and smaller model size, demonstrating
    superior data efficiency over existing world-model-based methods. World4Drive is marked separately as it employs an additional ViT-L depth estimator.}
    \label{fig:teaser}
\end{figure}

Recently, world-model-based approaches learn scene representations through temporal self-supervised learning,
providing dense supervision while reducing the dependence on perception labels. These methods can be broadly
categorized into two groups. The first group~\cite{zhang2025epona, li2026drivevla} learns driving planning through explicit video
generation. However, these methods suffer from high computational cost and tend to focus on visual details irrelevant to planning. The second group learns planning-relevant
representations through implicit future latent prediction~\cite{li2024law, zheng2025world4drive}. Although the latent representations
are relatively lightweight and capture dynamic information via temporal self-supervision, they still suffer from
insufficient representation quality. Specifically: (1) the latent representations remain inadequately compressed; (2)
they lack spatial understanding or rely on external depth estimation models at inference time, introducing additional
latency; and (3) historical and dynamic information are underutilized, as these methods are restricted to single-step transition ($T{\rightarrow}T{+}1$),
which severely limits the temporal understanding that driving demands. Autonomous driving requires modeling long-horizon interactions among surrounding agents. However, next-frame prediction provides only short-term supervision and misses the cumulative dynamics needed for robust planning. As illustrated in Fig.~\ref{fig:teaser}, these limitations lead to sub-optimal planning performance.

\begin{figure*}[t]
    \centering
    \includegraphics[width=0.9\linewidth]{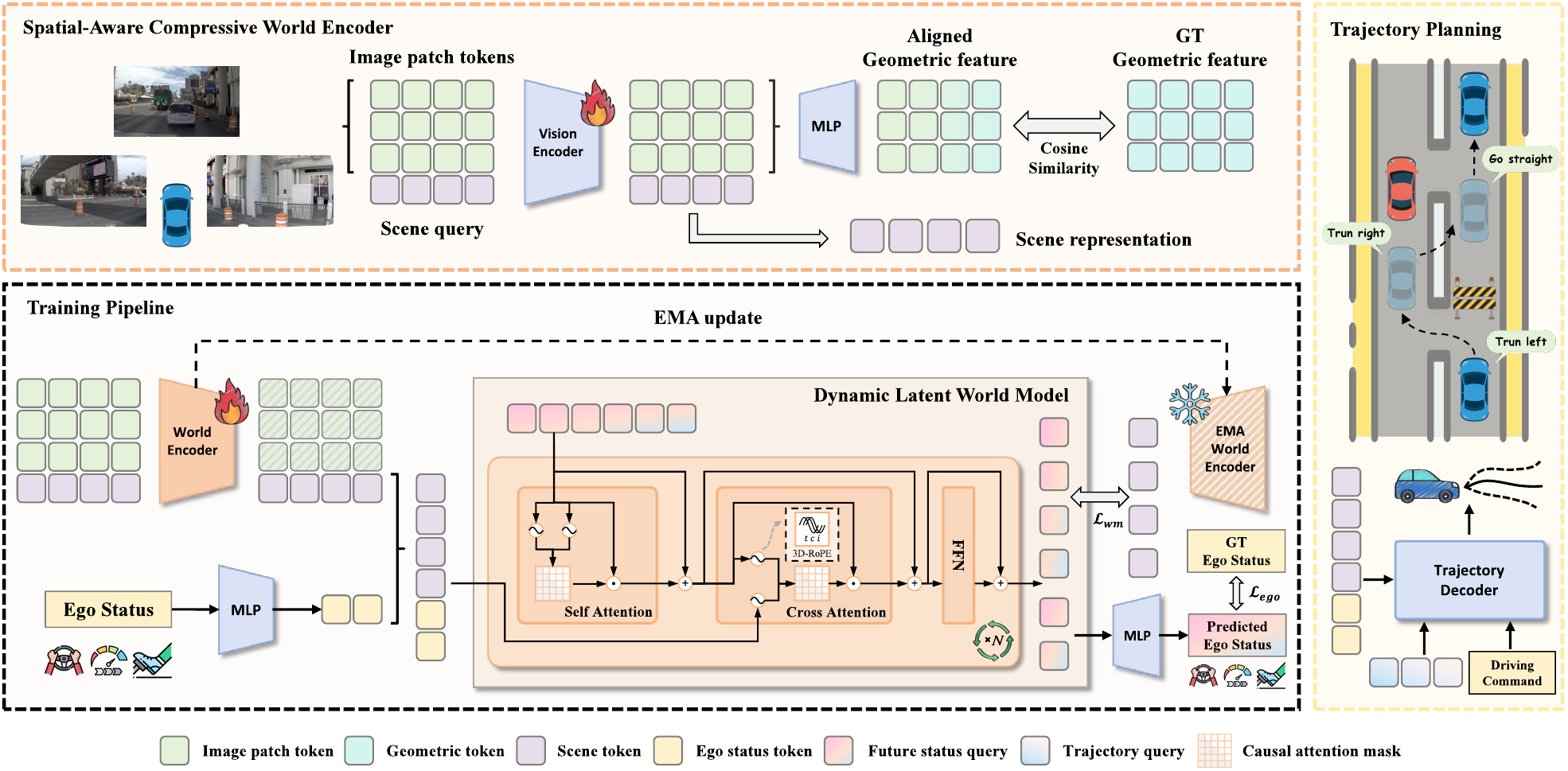}
    \caption{
        Overall architecture of Latent-WAM. The framework consists of three modules:
        (1) Spatial-Aware Compressive World Encoder (SCWE) compresses images into compact scene tokens with geometric distillation for spatial understanding;
        (2) Dynamic Latent World Model (DLWM) predicts future world states from historical scene representations and ego status;
        (3) Trajectory Decoder forecasts 4-second trajectories.
        SCWE and DLWM jointly optimize the vision backbone. Notably, DLWM is used only during training and discarded at inference, while its prediction objective enriches the vision backbone, enabling the trajectory decoder to directly use SCWE representations for planning.
    }
    \label{fig:architecture}
\end{figure*}

To address these issues, we propose Latent-WAM, which comprises a Spatial-Aware Compressive World Encoder (SCWE) and a
Dynamic Latent World Model (DLWM), targeting the two most planning-relevant understanding tasks—spatial and dynamic
understanding—to achieve highly compressed driving world representations with improved planning performance.
Specifically, SCWE distills knowledge from a geometric foundation model into the vision backbone and employs learnable
queries to extract spatially-informed tokens from the enriched features, achieving high compression of scene
information.
DLWM departs from prior single-step latent world models by autoregressively predicting \emph{multi-frame} future world status conditioned on historical scene and ego status representations. Rather than learning a local $T{\rightarrow}T{+}1$ transition, DLWM models the full trajectory of world evolution across multiple future timesteps via a causal Transformer, providing substantially richer self-supervised signals that encourage the encoder to capture long-range temporal dependencies and cumulative dynamics. Furthermore, DLWM jointly predicts both visual scene changes and ego status (driving command, velocity, acceleration), coupling scene dynamics with self-state awareness to produce representations inherently attuned to planning-relevant motion patterns. As a result, DLWM is entirely discarded at inference, with the lightweight trajectory decoder directly consuming the enriched representations to generate planned trajectories.

We extensively evaluate Latent-WAM on NAVSIM v2~\cite{cao2025navsimv2} and HUGSIM~\cite{zhou2024hugsim}. Latent-WAM achieves 89.3 EPDMS on
NAVSIM v2, 45.9 RC and 28.9 HD-score on HUGSIM, establishing new state-of-the-art results. Under the perception-free setting, our method outperforms previous approaches by 3.2 EPDMS. Notably, attention map visualization reveals that
our world representations are highly focused on spatial structures and driving intent, demonstrating the effectiveness
of our approach for planning-centric representation learning. Our contributions are summarized as follows:
\begin{itemize}
    \item We propose Spatial-Aware Compressive World Encoder, a novel scene compression method that distills geometric
    knowledge into the vision backbone to extract highly compressed, spatially-aware visual representations.
    \item We propose Dynamic Latent World Model, which autoregressively models multi-frame world evolution with a causal Transformer. Joint prediction of visual dynamics and ego status provides richer self-supervised signals for temporal understanding, producing planning-effective representations that eliminate the need for explicit future prediction at inference time.
    \item Our method achieves new state-of-the-art results on both NAVSIM v2 and HUGSIM, outperforming the best prior
    perception-free method~\cite{li2026drivevla} by 3.2 EPDMS.
\end{itemize}

\section{Related Work}
\subsection{Scene Representation in End-to-End Autonomous Driving}

End-to-end autonomous driving directly optimizes trajectory planning, making the intermediate scene representation critical to overall performance. Early works~\cite{hu2023uniad, transfuser} adopt dense
BEV layouts supervised by semantic maps and occupancy labels. Subsequent methods~\cite{jiang2023vad, chen2024vadv2, sun2024sparsedrive, jia2025drivetransformer} shift towards lightweight vectorized or
sparse representations for improved efficiency. Recent efforts leverage VLMs~\cite{tian2024token, pan2024vlp, Hegde2025DistillingML, zheng2026planagent} for richer semantic information, or employ latent world
models~\cite{li2024law, zheng2025world4drive} and video generation~\cite{xia2025drivelaw} for scene representation. Besides, recent end-to-end planners also explore generative trajectory modeling and safety-aware imitation learning. MeanFuser~\cite{wang2026meanfuser} achieves one-step multimodal trajectory generation, and Beyond Imitation~\cite{wang2026beyondimitation} enhances safety via expert demonstrations and hard negatives. However, these methods often exhibit weak 3D spatial understanding and rely on
cumbersome representations. We address these issues by distilling geometric knowledge into the vision backbone for stronger spatial understanding, while compressing scene information into a compact set
of tokens via learnable queries.

\subsection{World Models for Autonomous Driving}

World models have been widely adopted in autonomous driving for environment representation and future state prediction. One line of work applies video generation models~\cite{Hu2023GAIA1AG,
wang2023drivedreamer, zhao2024drivedreamer-2, chen2024driveworld, wang2023driving-wm, gao2024vista, guo2024dist4d} to construct pixel-level driving world representations, while 3D world models based on
occupancy and point clouds~\cite{zheng2023occworld, zhang2024copilotd, zyrianov2025lidardm, dang2025sparseworld} enforce geometric constraints in 3D space, with subsequent works~\cite{li2024uniscene,
li2025scaling} unifying 2D and 3D modeling. Building on these capabilities, some approaches~\cite{yang2024drivearena, li2025omninwm} use world models as closed-loop simulators, while
others~\cite{zhang2025epona, li2026drivevla, xia2025drivelaw, li2024law, zheng2025world4drive} directly leverage them for planning. To reduce computational overhead, LAW~\cite{li2024law},
World4Drive~\cite{zheng2025world4drive}, and Drive-JEPA~\cite{wang2026drivejepa} perform trajectory planning in the latent space. 
Recent works extend latent world models to reinforcement learning and safety-aware closed-loop planning: DreamerAD~\cite{yang2026dreamerad} leverages one for efficient policy optimization, while SoAD~\cite{gao2026soad} combines a vector world model with safety-oriented value estimation for trajectory evaluation.
Our method similarly adopts a latent world model, but fundamentally differs in two respects: (1)~we predict multi-frame future world status rather than the standard single-step transition, providing richer self-supervised signals for temporal understanding; and (2)~we jointly model scene dynamics and ego status, coupling spatial-temporal prediction with self-state awareness. We additionally incorporate 3D-RoPE for spatio-temporal position encoding.

\section{Method}

\subsection{Overall}
\label{sect:overall}

The overall architecture of Latent-WAM is illustrated in Fig.~\ref{fig:architecture}, consisting of three core modules: 1) Spatial-Aware Compressive World Encoder (Sec.~\ref{sect:encoder}), which uses learnable queries and a vision encoder to compress images into compact scene tokens and a geometric foundation model is used to distill geometric perception ability into the encoder to improve spatial understanding, 2) Dynamic Latent World Model (Sec.~\ref{sect:wm}), a causal transformer decoder that models world transition dynamics by
autoregressively predicting future world status conditioned on historical scene representations and ego status, and 3) Trajectory Decoder (Sec.~\ref{sect:traj}), which forecasts trajectories over a 4-second horizon from world status representations.
The SCWE and DLWM jointly optimize the vision backbone. Geometric distillation enhances spatial understanding, while self-supervised learning improves temporal dynamics modeling. Notably, the DLWM is used exclusively during training and is entirely discarded at inference time. The world prediction
objective serves to enrich the vision backbone with dynamic scene understanding, producing representations that already
encode the temporal foresight needed for planning. During inference, the trajectory decoder directly employs these
enriched representations from SCWE, eliminating any additional latency from auxiliary modules.

\subsection{Spatial-Aware Compressive World Encoder}
\label{sect:encoder}
To model long-horizon world state transition, world status representations need to be sufficiently lightweight while compressing rich scene information. Different from prior methods that rely on extensive visual information, we use only a small set of learnable queries that fully interact with image patch tokens from the inputs, compressing rich world perception information into the latent space. In addition, to enhance the spatial understanding of the vision backbone, we use a geometric foundation model WorldMirror~\cite{liu2025worldmirror} to inject geometric awareness into the backbone through distillation.
  
\subsubsection{\textbf{Scene Compression.}}
We use a set of compact scene tokens to compress the heavy vision tokens from multi-view images, forming a foundational component of world status representations.

Given sequential multi-view images $I \in \mathbb{R}^{\mathbf{T} \times \mathbf{M} \times H
\times W \times 3}$ as input, we first embed them into image patch tokens $X \in
\mathbb{R}^{\mathbf{T} \times \mathbf{M} \times \mathbf{S} \times D_e}$, where $\mathbf{T},
\mathbf{M}, \mathbf{S}$ denote the temporal horizon, number of cameras, and number of image
patches, respectively. A set of scene queries $Q_{\text{scene}} \in \mathbb{R}^{\mathbf{T} \times \mathbf{M} \times \mathbf{N} \times D_e}$ are randomly initialized and concatenated with $X$. The concatenated
features are fed into a DINO encoder $\mathcal{E}$ containing an MLP that projects to $D_l$-dimensional latent space, yielding frame-wise and
view-specific scene representations $\hat{Q}_{\text{scene}} \in \mathbb{R}^{\mathbf{T} \times
\mathbf{M} \times \mathbf{N} \times D_l}$ and image tokens $\hat{X} \in \mathbb{R}^{\mathbf{T}
\times \mathbf{M} \times \mathbf{S} \times D_l}$:
\begin{align}
    \hat{Q}_{\text{scene}}, \hat{X} = \mathcal{E} \left ( \left [ Q_{\text{scene}}; X\right ] \right )
\end{align}
where $\mathbf{N}$ is the number of scene query tokens, $D_e$ the encoder dimension, and $D_l$ the latent dimension.

By integrating scene queries with raw image tokens, extensive visual information from numerous image patch tokens is efficiently compressed into a compact set of tokens, significantly reducing computational overhead for subsequent long-term world model training and trajectory planning.

\subsubsection{\textbf{Geometric Alignment.}}
To distill the spatial understanding capabilities of geometric models into the DINO encoder, we employ the image patch tokens output by SCWE as carriers for receiving dense spatial-semantic information from geometric features. 

Multi-view images across consecutive frames $I$ are additionally fed into a geometric foundation model $f_g$, producing patch-level geometric features $f_g(I) \in \mathbb{R}^{\textbf{T} \times \textbf{M} \times \textbf{S} \times D_g}$. The DINO backbone outputs $\hat{X}$ are subsequently projected via a geometric projector $\phi$, yielding $\phi(\hat{X}) \in \mathbb{R}^{\textbf{T} \times \textbf{M} \times \textbf{S} \times D_g}$. Both $f_g(I)$ and $\phi(\hat{X})$ are first normalized using LayerNorm, then we compute their cosine similarity loss:
\begin{align}
    \mathcal{L}_{\text{align}} = 1 - \text{cos}\left(\text{LN}(\phi(\hat{X})), \text{LN}(f_g(I))\right)
\end{align}
where $\cos(\cdot, \cdot)$ denotes cosine similarity, $\text{LN}(\cdot)$ denotes LayerNorm.

Notably, since $f_g$ remains frozen throughout training, the geometric features can be pre-computed and offline-cached, enabling direct loading during training without incurring repeated inference costs or GPU memory overhead. This design substantially reduces training time and computational expenses.

\subsection{Dynamic Latent World Model}
\label{sect:wm}
In this section, we propose a Dynamic Latent World Model (DLWM). By predicting future world status representations causally in the latent space, the backbone obtains temporal dynamics modeling capabilities through a self-supervised training paradigm under the perception-free setting.

\subsubsection{\textbf{World Latent Status Aggregation.}}
The per-camera scene tokens $\hat{Q}_{\text{scene}}$ only capture isolated view-specific information, which is insufficient for holistic world modeling. To enable effective future prediction and trajectory planning, we aggregate these tokens and integrate with ego status to construct unified frame-wise world status representations.
The ego status across consecutive frames, which include driving commands, velocity, and acceleration, is encoded by an ego status encoder (a single-layer MLP) into ego status
embeddings $S_{\text{ego}} \in \mathbb{R}^{\textbf{T} \times D_l}$. 
The scene tokens $\hat{Q}_{\text{scene}}$ from the SCWE and the ego status embeddings $S_{\text{ego}}$ collectively constitute the frame-wise world latent state.

Specifically, scene tokens from different cameras are aggregated to form a holistic perception representation of the surrounding environment, which is subsequently concatenated with $S_{\text{ego}}$ encoding the ego status, yielding a unified Scene-Ego world status representation $S_{\text{world}} \in \mathbb{R}^{\textbf{T} \times (\textbf{M} \times \textbf{N} + 1)
\times D_l}$. 
This representation serves as the foundation for subsequent future world status prediction within the world model and trajectory planning.

\begin{figure}  %
    \centering
    \includegraphics[width=0.9\linewidth]{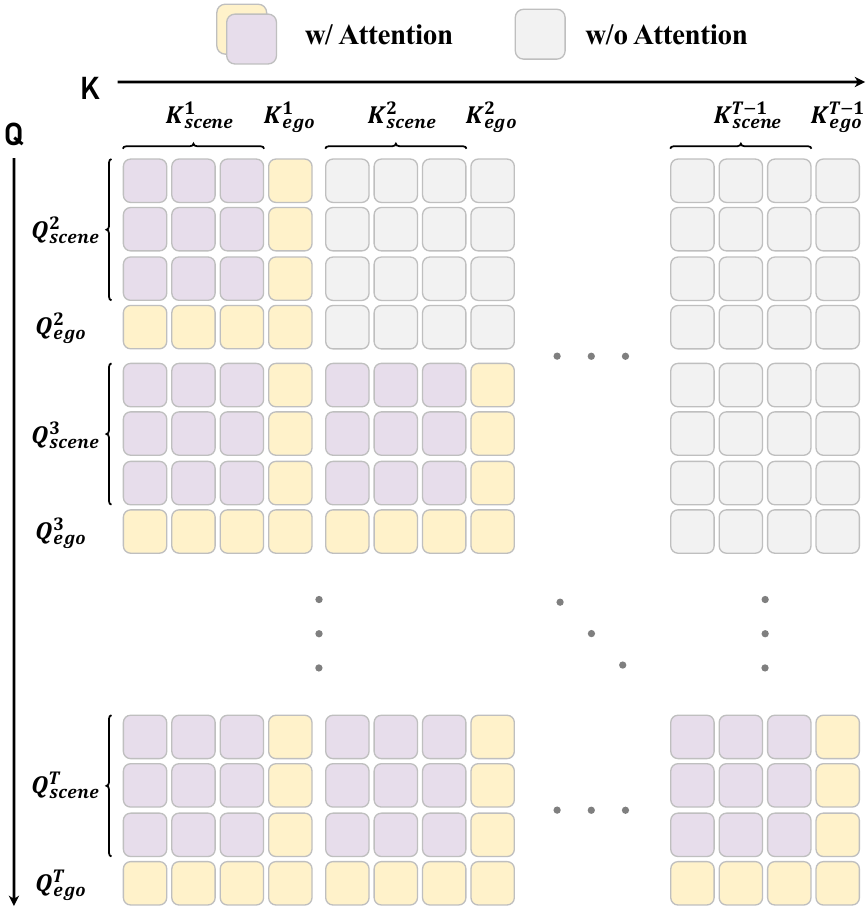}
    \caption{Teacher Forcing Attention Mask.}
    \label{fig:mask}
    \vspace{-1em}
\end{figure}

\subsubsection{\textbf{Causal World Model Prediction.}}
We formulate world state transition dynamics modeling as an autoregressive prediction problem, where all future frame world status predictions are conditioned on historical world status representations, encompassing holistic scene representations and ego status. This unified autoregressive framework enables a natural training strategy: treating the alternating Scene-Ego world status token sequence $S_{\text{world}}^i$ as frame-wise blocks and adopting the standard next-token prediction paradigm to train the world model.

In detail, we randomly initialize a set of learnable future world status queries $Q_{\text{future}} \in \mathbb{R}^{(\textbf{T}-1) \times (\textbf{M} \times \textbf{N} + 1) \times D_l}$, 
with 
$S_{\text{world}}^i, i \in \{1, \ldots,$ $T-1\}$ forming the key-value cache $KV_{\text{future}} \in \mathbb{R}^{(\textbf{T}-1) \times (\textbf{M} \times \textbf{N} + 1) \times D_l}$. 
The DLWM adopts a standard Transformer decoder incorporating Rotary Position Embedding, producing future world status predictions $S_{\text{future}} \in \mathbb{R}^{(\textbf{T}-1) \times (\textbf{M} \times \textbf{N} + 1) \times D_l}$. Formally, 
\begin{align}
    S_{\text{future}} = \text{DLWM}(Q_{\text{future}}, KV_{\text{future}})
\end{align}
The ground truth $S_{\text{future}}^{\text{GT}}$ is generated by a target encoder,
which is a frozen copy of the SCWE updated via Exponential Moving Average (EMA) to provide stable supervision
signals.

\subsubsection{\textbf{Teacher Forcing Attention Mask.}}
Given a Scene-Ego interleaved token sequence, the world model predicts each future token
by attending to all historical tokens. We adopt teacher forcing during training: ground
truth tokens serve as context for predicting subsequent tokens, preventing error accumulation across timesteps. 

Causal prediction is implemented via frame-wise attention masks, as shown in Fig.~\ref{fig:mask}.
Within each frame block, tokens attend bidirectionally to each other. Cross-frame, each
token can only attend to tokens appearing earlier in the sequence, enforcing temporal causality.
This frame-wise attention design enables parallel prediction of all future world status while
maintaining causal consistency, significantly improving training speed and efficiency.

\subsubsection{\textbf{3D-RoPE}}
Although $S_{\text{world}}$ contains world status representations across multiple timestamps and
camera views, the query vectors $Q_{\text{future}}$ and context features $KV_{\text{future}}$ do
not explicitly carry temporal or spatial position information. To enable the model to
distinguish temporal and spatial relationships among tokens, we inject spatio-temporal position
information into multi-head attention via 3D-RoPE, which differentiates tokens across timesteps,
  camera views, and positions within long sequences.

In practice, we split the head dimension $D_h$ into three parts to encode temporal coordinate $t$, camera index $m$, and token index $n$ into $Q_{\text{future}}$ and $KV_{\text{future}}$.
We use absolute position indices to encode all three coordinates: temporal index with frequency 50, camera index with frequency 10, and token index with frequency 100.

\subsubsection{\textbf{Ego Status Supervision.}}
The world status representation $S_{\text{world}}$ contains both scene tokens and ego
status embeddings. During world model prediction, ego status guides the evolution of scene
representations as well as itself. 
Therefore, accurate prediction of ego status is crucial for modeling world dynamics. To provide precise guidance for the transition of world state, we introduce ego status supervision built upon the self-supervised future status prediction. We extract ego status embeddings $S_{\text{ego}}^{i'}$ from the predicted future world status
$S_{\text{future}}$, where $i' \in \{2, \ldots, T\}$. These embeddings are then fed into three
separate MLPs: a driving command decoder $D_{\text{cmd}}$, a velocity decoder $D_v$, and an
acceleration decoder $D_a$, producing predictions $C_{\text{pred}} \in \mathbb{R}^{(\textbf{T}-1) \times
4}$, $V_{\text{pred}} \in \mathbb{R}^{(\textbf{T}-1) \times 2}$, and $A_{\text{pred}} \in \mathbb{R}^{(\textbf{T}-1) \times 2}$, respectively. 
Formally:
\begin{align}
  C_{\text{pred}} = \text{Softmax}(D_{\text{cmd}}(S_{\text{ego}}^{'})), \
  V_{\text{pred}} = D_v(S_{\text{ego}}^{'}), \
  A_{\text{pred}} = D_a(S_{\text{ego}}^{'}).
\end{align}
where $S_{\text{ego}}^{'} \in \mathbb{R}^{(\textbf{T}-1) \times D_l}$ is the set containing all predicted future ego status embeddings $S_{\text{ego}}^{i'}$ with $i’ \in \{2, \dots, T\}$.

\begin{table*}[t]
\centering
\caption{\textbf{Comparison with state-of-the-art methods on the NAVSIM v2 with extended metrics.} 
We indicate the best and second best with \textbf{bold} and \underline{underlined} respectively. $\dag$: The reported results are dependent on perception-based annotation.}
\label{tab:navsim_v2_results}
\resizebox{0.85\textwidth}{!}{
    \begin{tabular}{l|cccc|ccccc|>{\columncolor{gray!25}}c|c}
    \toprule
    \textbf{Method} & \textbf{NC$\uparrow$} & \textbf{DAC$\uparrow$} & \textbf{DDC$\uparrow$} & \textbf{TLC$\uparrow$} & \textbf{EP$\uparrow$} & \textbf{TTC$\uparrow$} & \textbf{LK$\uparrow$} & \textbf{HC$\uparrow$} & \textbf{EC$\uparrow$} & \textbf{EPDMS$\uparrow$} & \textbf{Latency (s)} $\downarrow$\\
    \midrule
    \multicolumn{9}{l}{\textit{Perception-based Methods}} \\
    \cellcolor{blue!10}TransFuser~\cite{transfuser}            
                                        & 96.9              & 89.9              & 97.8              & \underline{99.7}  & 87.1              & 95.4              & 92.7              & \textbf{98.3}     & 87.2              & 76.7              & 0.02s              \\
    \cellcolor{blue!10}DriveSuprim~\cite{yao2025drivesuprim}   
                                        & 97.5              & 96.5              & 99.4              & 99.6              & 88.4              & 96.6              & 95.5              & \textbf{98.3}     & 77.0              & 83.1              & ---               \\
    \cellcolor{blue!10}ReCogDrive~\cite{xiong2026recogdrive}
                                        & 98.3              & 95.2              & \underline{99.5}  & \textbf{99.8}     & 87.1              & 97.5              & 96.6              & \textbf{98.3}     & 86.5              & 83.6              & ---               \\
    \cellcolor{blue!10}DiffusionDrive~\cite{diffdrive}
                                        & 98.2              & 95.9              & 99.4              & \textbf{99.8}     & 87.5              & 97.3              & 96.8              & \textbf{98.3}     & \textbf{87.7}     & 84.5              & 0.03s              \\
    \cellcolor{blue!10}WorldRFT\cite{yang2026worldrft}     
                                        & 97.8              & 96.5              & \underline{99.5} 
                                        & \textbf{99.8}     & \underline{88.5}  & 97.0 
                                        & 97.4              & \underline{98.1}  & 69.1 
                                        & 86.7              & 0.29s             \\
    \cellcolor{blue!10}GoalFlow\cite{xing2025goalflow}
                                        & \underline{98.4}  & 98.2              & \underline{99.5} 
                                        & \textbf{99.8}     & 88.1              & 97.4
                                        & 96.7              & \underline{98.1}              & 57.9 
                                        & 87.3              & ---             \\
    \cellcolor{blue!10}Drive-JEPA$^{\dag}$~\cite{wang2026drivejepa}
                                        & \underline{98.4}  & \underline{98.6}  & 99.1              & \textbf{99.8}     & 88.4              & \underline{97.8}  & \underline{97.6}  & 97.9              & 84.8              & \underline{87.8}  & ---               \\
    \midrule
    \multicolumn{9}{l}{\textit{Perception-free Methods}} \\
    \cellcolor{yellow!10}World4Drive~\cite{zheng2025world4drive}
                                        & 97.8              & 96.3              & 99.4              & \textbf{99.8}     & 88.3              & 97.1              & \textbf{97.7}     & 98.0              & 53.9              & 84.8              & 0.73s             \\ 
    \cellcolor{yellow!10}Epona ~\cite{zhang2025epona}    
                                        & 97.1              & 95.7              & 99.3 
                                        & \underline{99.7}  & \textbf{88.6}     & 96.3 
                                        & 97.0              & 98.0              & 67.8 
                                        & 85.1              & 2.00s               \\
    \cellcolor{yellow!10}DriveVLA-W0~\cite{li2026drivevla}
                                        & \textbf{98.5}     & \textbf{99.1}     & 98.0              & \underline{99.7}  & 86.4              & \textbf{98.1}     & 93.2              & 97.9              & 58.9              & 86.1              & ---               \\
    \midrule
    \cellcolor{yellow!10}Latent-WAM (Ours)
                                        & 98.1              & 97.3              & \textbf{99.6}     & \textbf{99.8}     & 87.7              & 97.3              & \underline{97.6}  & \underline{98.1}  & \underline{87.3}  & \textbf{89.3}     & 0.11s             \\
    \bottomrule
    \end{tabular}
}
\end{table*}

\subsection{Trajectory Planning} 
\label{sect:traj}
We employ a trajectory decoder $D_{\tau}$ for planning, which generates trajectories corresponding to different driving intentions based on the control commands.
A set of randomly initialized learnable trajectory queries $Q^t_{\tau} \in \mathbb{R}^{K \times n_p \times D_l}$ is fed into the trajectory decoder along with the current world status representation $S_{world}^t$ of the driving scenario. 
After processing, the trajectory tokens are decoded via a lightweight MLP into $K$ candidate trajectories. 
Finally, conditioned on the driving command $C^t$ at the current timestep $t$, the corresponding candidate trajectory is selected as the final trajectory $\tau$. Formally, 
\begin{align}
    \tau = D_{\tau}(Q^t_{\tau}, S_{world}^t, C^t)
\end{align}
Each decoded candidate trajectory $\tau_i$ is a sequence of $n_p$ poses spanning from the current timestep $t$ to a future time $t + T$, where $T$ represents the total prediction horizon. Each pose is represented as  $(x, y, \theta) \in \mathbb{R}^3$, yielding complete trajectories in $\mathbb{R}^{n_p \times 3}$, where $x$ and $y$ correspond to longitudinal and lateral displacements, respectively, with $\theta$ denoting the heading angle. All quantities are referenced to the ego vehicle's local coordinate system at timestep $t$. The time interval between consecutive predicted poses is assumed to be uniform.

\subsection{Training and Inference}
\textbf{Training Objective.}
Following prior work, we employ $L_1$ loss $\mathcal{L}_{traj}$ to optimize the generated multi-candidate trajectories by imitating expert trajectories. Geometric alignment computes cosine similarity loss $\mathcal{L}_{align}$ on normalized features, maximizing cosine similarity to distill spatial understanding capability. For the world model, we use MSE loss $\mathcal{L}_{wm}$ to optimize the prediction of future world state. Future ego status requires supervision over driving command, velocity, and acceleration, including 
$\mathcal{L}_{cmd} = \text{CrossEntropy}(C_{pred}, C_{gt})$, $\mathcal{L}_v = \text{MSE}(V_{pred}, V_{gt})$ and $\mathcal{L}_a = \text{MSE}$ \\ $(A_{pred}, A_{gt})$. The total loss for ego status is $\mathcal{L}_{ego} = \mathcal{L}_{cmd} + \mathcal{L}_v + \mathcal{L}_a$. The final loss for end-to-end training is: 
\begin{align}
    \mathcal{L} = \mathcal{L}_{traj} + \alpha \mathcal{L}_{align} + \beta \mathcal{L}_{wm} + \gamma \mathcal{L}_{ego}
\end{align}
where $\alpha = 0.1$, $\beta = 0.2$, and $\gamma = 0.1$. \\
\textbf{Inference.}
During inference, the Dynamic Latent World Model is entirely bypassed---only the Spatial-Aware Compressive World
Encoder and the Trajectory Decoder are required. Since the training objective already endows the scene representations
with temporal dynamics and ego-awareness, no explicit future prediction is needed at test time. Inference reduces to a
single forward pass through the encoder and decoder, without any auxiliary modules.

\section{Experiment}
\subsection{Implementation Details}
\subsubsection{\textbf{Architecture.}}
We adopt DINOv2-Base \cite{oquab2023dinov2} as our vision encoder (86.6M parameters), which
constitutes the foundation of the Spatial-Aware Compressive World Encoder (SCWE) and produces patch tokens of dimension
$D_e=768$. We employ $N=16$ learnable scene queries with dimension $D_e$, projected to latent
space $D_l=256$ via an MLP. We set the temporal horizon to $T=4$ frames and adopt three camera views—left, front, and right. Each view is resized to $224 \times 448$ before feeding into the World Encoder. Notably, the geometric foundation model we employ is WorldMirror \cite{liu2025worldmirror}, a feed-forward model built upon VGGT \cite{wang2025vggt}. The dimension of ground-truth geometric feature map is $D_g=2048$.
For the DLWM, we design a causal transformer decoder with 3D-RoPE for position embedding. The trajectory decoder follows a standard transformer architecture. Both DLWM and trajectory decoder comprise 4 layers with 8 attention heads, hidden dimension 256, and FFN dimension 1024. To provide ground truth world status representations for self-supervised training, we additionally introduce a frozen SCWE updated via Exponential Moving Average (EMA) with momentum 0.999, maintaining complete architectural symmetry with the online encoder.
The model contains 104M parameters at inference time. During training, an additional EMA encoder is introduced for self-supervised learning, bringing the total to 191M, of which only 104M are trainable.

\subsubsection{\textbf{Training Configuration.}} 
Latent-WAM is trained on 32 A100 GPUs for 100 epochs with a total batch size of 512, taking about two days. We use AdamW with learning rate $2 \times 10^{-4}$ and weight decay 0.05, with linear warm-up over the first 10\% steps followed by cosine annealing to $1 \times 10^{-6}$. BF16 mixed-precision training is applied throughout.

\begin{table}[]
    \centering
    \caption{\textbf{Photorealistic closed-loop evaluation on HUGSIM \cite{zhou2024hugsim}}. Zero-shot generalization using our model trained from NAVSIM-v2. Scores are per difficulty and overall average road completion (RC) and HD-Score, higher always better.}
    \label{tab:hugsim}
    \setlength{\tabcolsep}{3pt}
    \resizebox{0.475\textwidth}{!}{
        \begin{tabular}{l|ccccc|ccccc}
            \toprule
            \multirow{2}{*}{\textbf{Method}} &
            \multicolumn{5}{c|}{\textbf{RC$\uparrow$}} &
            \multicolumn{5}{c}{\textbf{HD-Score$\uparrow$}} \\
            & E & M & H & X & \textbf{Avg}. & E & M & H & X & \textbf{Avg}. \\
            \midrule
            UniAD\cite{hu2023uniad} &
            58.6 & 41.2 & 40.4 & 26.0 & 40.6 &
            48.7 & 29.5 & 27.3 & 14.3 & \textbf{28.9} \\
            VAD\cite{jiang2023vad} &
            38.7 & 27.0 & 25.5 & 23.0 & 27.9 &
            24.3 & \phantom{0}9.9 & 10.4 & \phantom{0}8.2 & 12.3 \\
            LTF\cite{chitta2022transfuser} &
            68.4 & 40.7 & 36.9 & 25.5 & 41.4 &
            52.8 & 24.6 & 19.8 & \phantom{0}8.1 & 24.8 \\
            GTRS-Dense\cite{li2025gtrs} &
            64.2 & 50.0 & 20.7 & 22.3 & 38.0 &
            55.5 & 39.0 & 11.7 & 14.3 & 28.6 \\
            \midrule
            Latent-WAM &
            84.2 & 42.5 & 30.6 & 35.5 & \textbf{45.9} &
            72.5 & 24.0 & 12.2 & 18.1 & \textbf{28.9} \\
            \bottomrule
            \multicolumn{11}{r}{\textit{\small E: Easy, M: Medium, H: Hard, X: Extreme}}
        \end{tabular}
    }
\end{table}

\subsection{Benchmarks \& Main Results}
\subsubsection{\textbf{NAVSIM}}
NAVSIM~\cite{im2024navsim} is a real-world dataset built upon OpenScene~\cite{openscene2023} and nuPlan~\cite{nuplan}, containing 103k training and 12k evaluation scenarios. NAVSIM v1 evaluates closed-loop planning via PDMS, which aggregates No at-fault Collisions (NC), Drivable Area Compliance (DAC), Time to Collision (TTC), Ego Progress (EP), and Comfort (C). NAVSIM v2~\cite{cao2025navsimv2} extends PDMS to EPDMS by adding Driving Direction Compliance (DDC), Traffic Light Compliance (TLC), Lane Keeping (LK), and refining Comfort into History Comfort (HC) and Extended Comfort (EC).

On NAVSIM v2 (Table~\ref{tab:navsim_v2_results}), Latent-WAM achieves the highest EPDMS among all methods, including those using perception annotations, and ranks top in DDC, TLC, and LK. Compared to Drive-JEPA~\cite{wang2026drivejepa}, which relies on perception annotations, our perception-free approach is slightly behind on safety metrics but substantially better on EC.

\subsubsection{\textbf{HUGSIM}}
For closed-loop evaluation, we employ HUGSIM~\cite{zhou2024hugsim}, which reconstructs scenarios from KITTI-360~\cite{kitti360}, nuScenes~\cite{nuscenes}, PandaSet~\cite{pandaset}, and Waymo~\cite{waymo} as photorealistic 3D environments, where the planner controls the ego vehicle via RGB cameras with dynamically adjusted viewpoints. On its pre-challenge test set (436 scenarios, four difficulty levels), we adopt the zero-shot protocol and report Road Completion (RC) and HUGSIM Driving Score (HD-Score), with HD-Score combining RC and averaged NC, DAC, TTC, and comfort. Latent-WAM, trained only on NAVSIM without fine-tuning, achieves 45.9 RC and 28.9 HD-Score, ranking first in RC and matching the best HD-Score among all baselines.

\subsubsection{\textbf{Inference Latency}}
As illustrated in Table~\ref{tab:navsim_v2_results}, we report the inference latency of each method on an A100. 
Our Latent-WAM has a total latency of 107 ms, meeting the requirements for real-time deployment.

\subsection{Ablation Study}
\subsubsection{\textbf{Effectiveness of Each Component.}}
We conduct progressive ablation experiments by gradually adding modules to the baseline, as shown in
Table~\ref{tab:ablation_all}.

\textbf{Scene Representation.}
The baseline directly feeds image patch tokens to the trajectory decoder, achieving 87.9 EPDMS. To enable long-horizon prediction in the world model, we compress image patches into compact scene tokens,
incurring
only a negligible performance drop ($\downarrow0.2$).
Injecting geometric information boosts performance to 88.6, demonstrating that geometric perception is crucial for
accurate trajectory planning.

\textbf{Dynamic World Modeling.}
Building upon compressed scene tokens, the DLWM improves performance from 87.7 to 88.0 by capturing
future dynamics. Adding ego status further increases the score to 88.3, indicating that self-state awareness benefits
planning. With geometric information, the world model further reaches 89.0. The full model combining all components achieves 89.3, showing that all modules contribute synergistically.

\begin{table}[t]
    \centering
    \caption{\textbf{Ablation study of each proposed component. \textcolor{green!60!black}{+} and \textcolor{red}{-} denote improvement/degradation relative to the baseline.}}
    \label{tab:ablation_all}
    \resizebox{0.475\textwidth}{!}
    {
        \begin{tabular}{cc|cc|c}
            \toprule
            \multicolumn{2}{c|}{\textbf{Scene Representation}} & \multicolumn{2}{c|}{\ \textbf{Dynamic World Modeling} \ } & \multirow{2}{*}[-0.75ex]{\ \textbf{EPDMS}$\uparrow$} \\
            \cmidrule{1-2} \cmidrule{3-4}
            \ Compression & \quad Geometry \ & \ World Model & \quad Ego Status \ &  \\
            \midrule
            \XSolidBrush & \XSolidBrush & \XSolidBrush & \XSolidBrush & 87.9 \\
            \Checkmark & \XSolidBrush & \XSolidBrush & \XSolidBrush & 87.7\down{0.2} \\
            \Checkmark & \XSolidBrush & \Checkmark & \XSolidBrush & 88.0\up{0.1} \\
            \Checkmark & \XSolidBrush & \Checkmark & \Checkmark & 88.3\up{0.4} \\
            \Checkmark & \Checkmark & \XSolidBrush & \XSolidBrush & 88.6\up{0.7} \\
            \Checkmark & \Checkmark & \Checkmark & \XSolidBrush & 89.0\up{1.1} \\
            \Checkmark & \Checkmark & \Checkmark & \Checkmark & \textbf{89.3}\up{1.4} \\
            \bottomrule
        \end{tabular}
    }
\end{table}
\begin{table}[t]
    \centering
    \caption{\textbf{Ablation study on geometry injection method}}
        \begin{tabular}{c|c}
            \toprule
            \textbf{Geometric Information} \ &\textbf{\ EPDMS$\uparrow$} \\
            \midrule
            w/o Geometric feature \ & 88.3 \\
            Concatenation & 88.0 \\
            Distillation & \textbf{89.3} \\
            \bottomrule
            \end{tabular}
        \label{tab:ablation_geom}
\end{table}

\subsubsection{\textbf{Impact of Geometric Information.}}
We compare geometric information integration approaches in Table~\ref{tab:ablation_geom}. Without geometry, the full model achieves 88.3 EPDMS. Directly concatenating frozen geometry features as key-value inputs degrades performance to 88.0, likely due to misalignment with the planning objective. In contrast, distilling geometric knowledge into the vision backbone improves EPDMS to 89.3, as end-to-end fine-tuning learns spatial-aware representations inherently aligned with downstream planning.

\subsubsection{\textbf{Vision Backbone for Geometric Distillation.}}
We compare different backbone scales and fine-tuning strategies for geometric distillation, as shown in
Table~\ref{tab:ablation_backbone}.

\begin{figure*}[htbp]
    \centering
    \includegraphics[width=\linewidth]{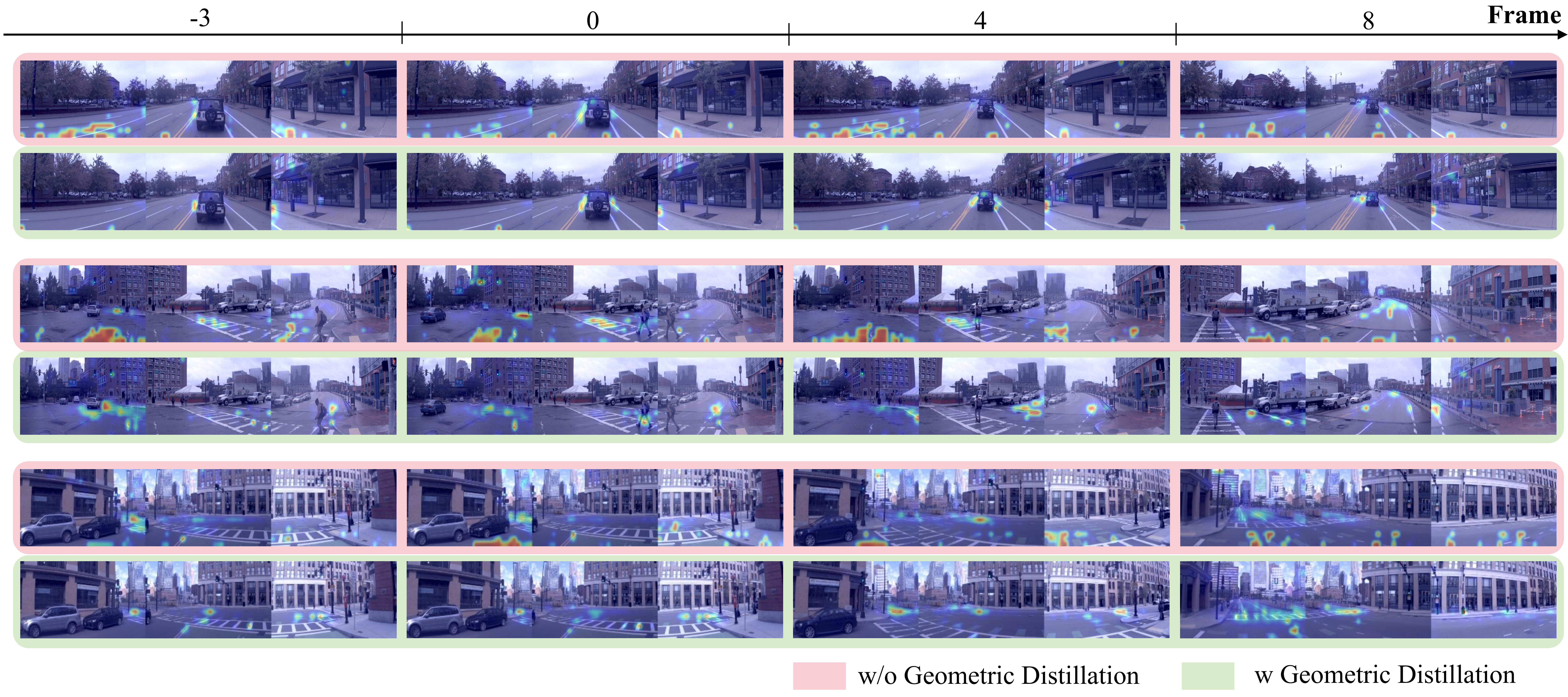}
    \caption{Visualization of attention maps between scene tokens and image patches, comparing our SCWE (with geometric distillation) against a baseline (without distillation). Our model exhibits three key properties: \textbf{(1)}~More focused attention distributions that concentrate on lane markings and drivable areas rather than irrelevant backgrounds; \textbf{(2)}~Stronger alignment with geometric structure, producing compact patterns that tightly follow lane boundaries and obstacle contours; \textbf{(3)}~Intent-aware attention that dynamically shifts focus toward regions along the intended driving direction.}
    \label{fig:attention_map_vis}
\end{figure*}

\textbf{Backbone Scale.}
DINO-Small achieves 86.3 EPDMS but remains suboptimal due to insufficient parameter capacity for high-dimensional
geometric features. DINO-Base with full fine-tuning achieves the best performance (89.3), indicating that sufficient
backbone capacity is essential for effective distillation.

\textbf{Training Strategy.}
Parameter-efficient fine-tuning via LoRA leads to severe degradation and unstable training for both DINO-Small-LoRA
(84.7) and DINO-Base-LoRA (68.5). LoRA's low-rank constraints are inadequate for distilling high-dimensional geometric
features, which require full parameter updates to capture geometric awareness. Notably, DINO-Base-LoRA degrades more severely than DINO-Small-LoRA, as the larger model has more parameters frozen under LoRA, amplifying the mismatch
between the low-rank update subspace and the high-dimensional distillation target.

In summary, effective geometric distillation requires both sufficient model capacity and full fine-tuning flexibility.
\begin{table}[t]
    \centering
    \caption{\textbf{Ablation study on vision backbone}}
    \label{tab:ablation_backbone}
        \begin{tabular}{c|c}
        \toprule
        Dino Backbone &\textbf{EPDMS$\uparrow$} \\
        \midrule
        Small & 86.3 \\
        Base & \textbf{89.3} \\
        Small-LoRA & 84.7 \\
        Base-LoRA & 68.5 \\
        \bottomrule
        \end{tabular}
\end{table}
\begin{table}[t]
    \centering
    \caption{\textbf{Ablation study on world model prediction stride}}
    \label{tab:ablation_step}
    \begin{tabular}{c|c}
    \toprule
    \textbf{Prediction Temporal Stride} \ & \ \textbf{EPDMS$\uparrow$} \\
    \midrule
    $0 \rightarrow 8$ & 88.4 \\
    $-3 \rightarrow 0 \rightarrow 4 \rightarrow 8$ & \textbf{89.3} \\
    $-3 \rightarrow -2 \rightarrow -1 \rightarrow 0 \rightarrow 2 \rightarrow 4 \rightarrow 6 \rightarrow 8$ \ & 89.1 \\
    \bottomrule
    \end{tabular}
\end{table}

\subsubsection{\textbf{World Model Prediction Temporal Stride.}}
We investigate the effect of prediction temporal stride in the DLWM (Table~\ref{tab:ablation_step}). Predicting only the final frame ($0 \rightarrow 8$) achieves 88.4 EPDMS. Incorporating historical and intermediate future predictions with stride 4 ($-3 \rightarrow 0 \rightarrow 4 \rightarrow 8$) improves performance to 89.3, as multi-step prediction provides richer supervision. Further densifying to stride 2 ($-3 \rightarrow -2 \rightarrow -1 \rightarrow 0 \rightarrow 2 \rightarrow 4 \rightarrow 6 \rightarrow 8$) yields 89.1, providing no additional gain. Since all configurations predict to the same horizon, denser sampling does not extend temporal reasoning; adjacent frames are highly similar and offer limited new signal, while more prediction targets increase optimization burden and computational cost. Thus, moderate prediction density best balances supervision effectiveness and training efficiency.

\subsection{\textbf{Qualitative Analysis.}}
\label{sec:qualitative}
We qualitatively evaluate Latent-WAM on NAVSIM, focusing on the spatial understanding of the World Encoder.

\begin{figure}[t]
  \centering
  \includegraphics[width=\linewidth]{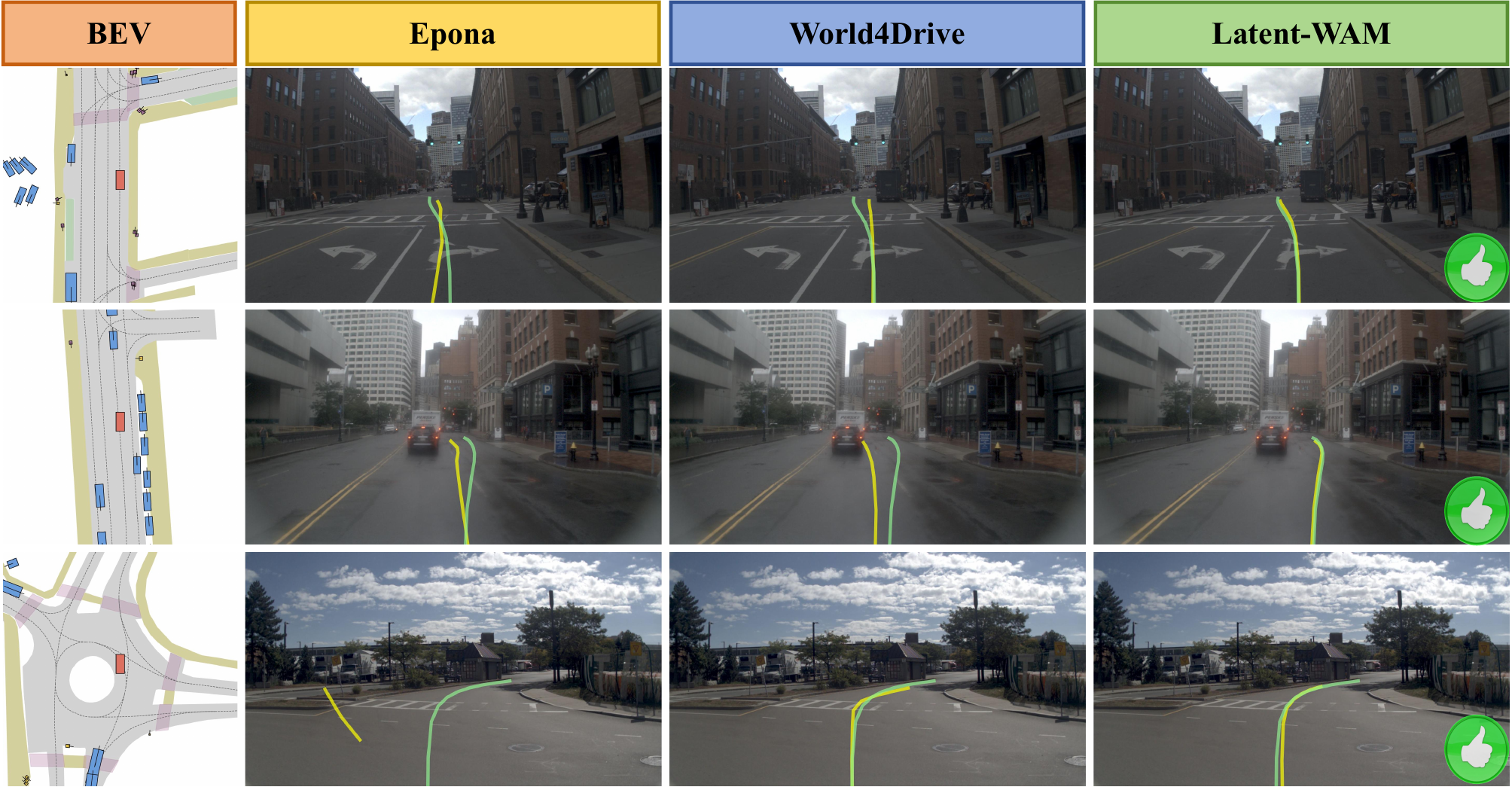}
  \caption{Visualization of planning trajectories. Green: human expert demonstration. Yellow: predicted trajectory of the corresponding method.}
  \label{fig:traj_vis}
\end{figure}

\textbf{Spatial Understanding.}
To investigate how geometric distillation enhances spatial understanding, we visualize attention maps between scene tokens and image patches (Fig.~\ref{fig:attention_map_vis}). Compared to the baseline without geometric distillation, our model exhibits much more focused attention, concentrating on lane markings, scene structures, and drivable areas critical for trajectory planning; the baseline distributes attention broadly across irrelevant background regions, suggesting that without geometric supervision, compressed scene tokens encode noisy information that interferes with planning.

Our attention maps also align more strongly with geometric structure: the baseline shows loose, diffuse regions, while our World Encoder produces compact patterns that tightly follow boundaries such as lane markings and open spaces near obstacles. Under different driving intentions—going straight, turning right, and turning left—attention strongly correlates with the intended direction, with minimal attention given to regions deviating from the planned trajectory. This intent-aware behavior demonstrates that our World Encoder selectively focuses on task-relevant spatial information conditioned on the driving context.

\textbf{Trajectory Comparison.}
Fig.~\ref{fig:traj_vis} compares trajectory predictions across methods. Our method better matches human trajectories and maintains safer distances from vehicles, unlike Epona~\cite{zhang2025epona} and World4Drive~\cite{zheng2025world4drive}, which yield sub-optimal paths.

\section{Conclusion}
In this paper, we presented Latent-WAM, an end-to-end autonomous driving framework that builds compact, planning-relevant world representations using a Spatial-Aware Compressive World Encoder to distill geometry while compressing multi-view images into compact scene tokens, and a Dynamic Latent World Model that autoregressively predicts future world states during training to equip the encoder with dynamic scene understanding. DLWM is discarded at inference, yet its prediction objective substantially improves planning performance, indicating that the value of latent world modeling lies in representation learning rather than explicit future prediction. At inference, only SCWE and a lightweight trajectory decoder are needed, yielding a compact 104M-parameter model that runs in a single forward pass. Latent-WAM achieves state-of-the-art results on NAVSIM v2 (89.3 EPDMS) and HUGSIM (45.9 RC and 28.9 HD-Score in zero-shot evaluation), outperforming the best prior perception-free method by 3.2 EPDMS with significantly less training data and fewer parameters. Qualitative analysis reveals focused, geometry-aligned attention patterns, validating the effectiveness of combining spatial-aware compression with dynamic world modeling.

\clearpage

\bibliographystyle{ACM-Reference-Format}
\bibliography{sample-base}

\balance
\clearpage
\nobalance
\appendix
\section*{Supplementary Material}
\section{Metrics}                                                      
\subsection{NAVSIM}                                                                                                                   
NAVSIM v2 \cite{cao2025navsimv2} extends the PDMS metric from NAVSIM v1 \cite{im2024navsim} to the Extended PDMS (EPDMS):                                                              
\begin{equation}
  \begin{split}
  \text{EPDMS} = &\ \text{NC} \times \text{DAC} \times \text{DDC} \times \text{TLC} \\
                  &\ \times \frac{5 \times (\text{EP} + \text{TTC}) + 2 \times (\text{LK} + \text{HC} + \text{EC})}{16}
  \end{split}
\end{equation}            
where NC, DAC, DDC, TLC, EP, TTC, LK, HC, and EC denote No at-fault Collision, Drivable Area Compliance, Driving Direction Compliance,      
Traffic Light Compliance, Ego Progress, Time to Collision, Lane Keeping, History Comfort, and Extended Comfort, respectively. Among these,  
DDC, TLC, LK, HC, and EC are newly introduced in NAVSIM v2.

\subsection{HUGSIM}
The HUGSIM \cite{zhou2024hugsim} Driving Score (HD-Score) at timestamp $t$ is computed as the product of driving policy items and a weighted average of           
contributory items:
\begin{equation}
\text{HD-Score}_t = \text{NC} \times \text{DAC} \times \frac{5 \times \text{TTC} + 2 \times \text{COM}}{7}                                  
\end{equation}
where NC, DAC, TTC, and COM denote No at-fault Collision, Drivable Area Compliance, Time to Collision, and Ego Comfort, respectively. These metrics follow the same definitions as used in NAVSIM v1.

The final HD-Score averages per-frame scores across all timestamps and scales by the route completion score $R_c \in [0,1]$:                
\begin{equation}
\text{HD-Score} = R_c \times \frac{1}{T}\sum_{t=0}^{T} \text{HD-Score}_t
\end{equation}

Notably, NC and TTC account for collisions with static background entities (e.g., buildings, fences, vegetation) using semantic             
segmentation.

\begin{figure*}  %
    \centering
    \includegraphics[width=0.96\linewidth]{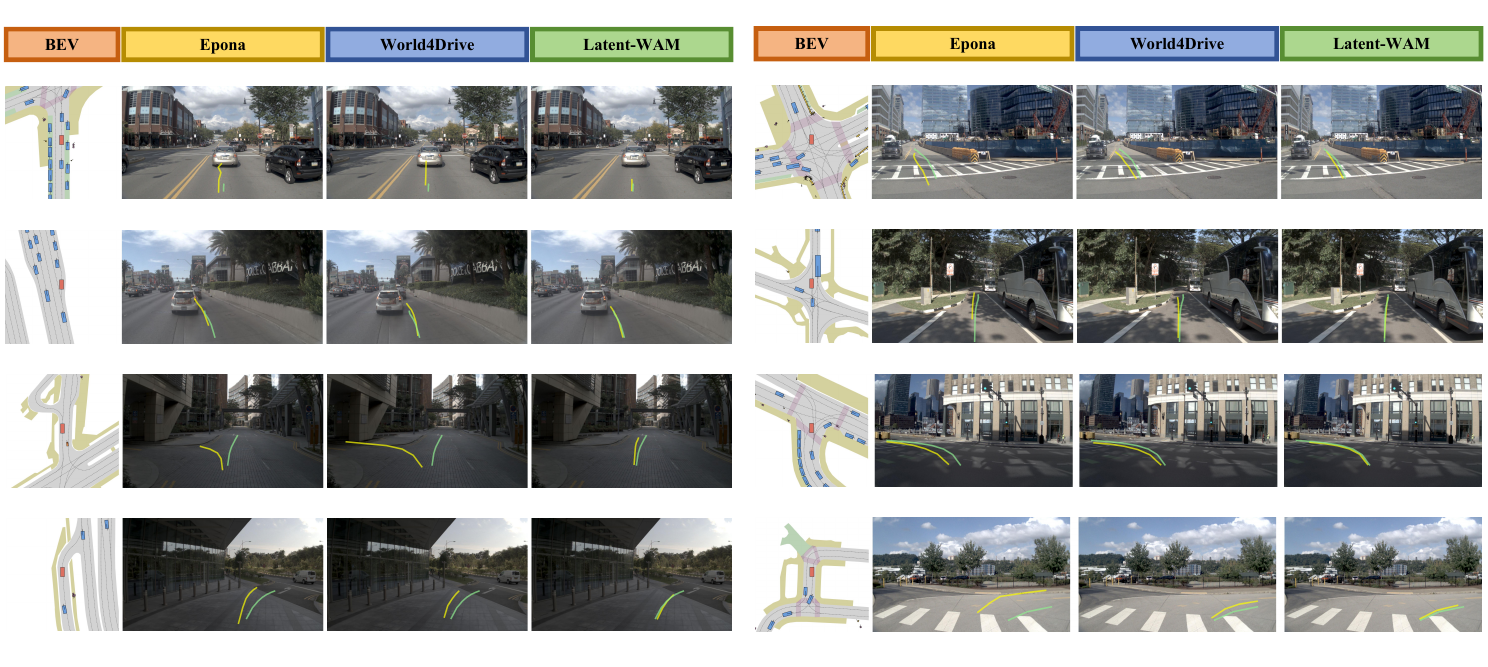}
    \captionof{figure}{Additional trajectory visualizations on NAVSIM. Green: human driving trajectory. Yellow: predicted trajectory from the corresponding method.}
    \label{fig:traj_append_1}
\end{figure*}

\section{Data Processing Pipeline}
\subsection{Image Preprocessing}

We utilize three camera views (left, front, right) as input. Each image undergoes a two-stage preprocessing pipeline: resizing followed by center cropping. Specifically, the original $1920 \times     
1080$ images are first resized to $455 \times 256$, then center-cropped to the final resolution of $448 \times 224$, yielding an aspect ratio of 2:1.                                                   
\subsection{Camera Intrinsic Adjustment}
When applying geometric feature extraction to images, the camera intrinsic matrix must be adjusted accordingly to maintain geometric consistency. Given the original intrinsic matrix $\mathbf{K}$:
\begin{equation}
    \mathbf{K} = \begin{bmatrix}
        f_x &   0 & c_x \\
          0 & f_y & c_y \\
          0 &   0 &   1
    \end{bmatrix},
\end{equation}
where $f_x$ and $f_y$ denote the focal lengths along the $x$ and $y$ axes, and $(c_x, c_y)$ represents the principal point.
After resizing from $(W_{\text{orig}}, H_{\text{orig}})$ to $(W_{\text{resize}}, H_{\text{resize}})$, the scale factors are computed as:
\begin{equation}
    s_x = \frac{W_{\text{resize}}}{W_{\text{orig}}}, \quad s_y = \frac{H_{\text{resize}}}{H_{\text{orig}}}.
\end{equation}
For center cropping to $(W_{\text{crop}}, H_{\text{crop}})$, the crop offsets are:
\begin{equation}
    \Delta_x = \frac{W_{\text{resize}} - W_{\text{crop}}}{2}, \quad \Delta_y = \frac{H_{\text{resize}} - H_{\text{crop}}}{2}.
\end{equation}
The adjusted intrinsic parameters are then:
\begin{equation}
    f'_x = f_x \cdot s_x, \quad
    f'_y = f_y \cdot s_y, \quad
    c'_x = c_x \cdot s_x - \Delta_x, \quad
    c'_y = c_y \cdot s_y - \Delta_y,
\end{equation}
yielding the adjusted intrinsic matrix $\mathbf{K}'$:
\begin{equation}
    \mathbf{K}' = \begin{bmatrix}
        f'_x & 0    & c'_x \\
        0    & f'_y & c'_y \\
        0    & 0    & 1
    \end{bmatrix}.
\end{equation}
\subsection{Camera Extrinsic Transformation}
The camera extrinsic matrix represents the transformation from camera coordinates to world (LiDAR) coordinates. We construct the camera-to-world matrix $\mathbf{T}_{c \to w} \in \mathbb{R}^{4 \times  
4}$ as:
\begin{equation}
    \mathbf{T}_{c \to w} = 
    \begin{bmatrix}
        \mathbf{R}_{c \to w} & \mathbf{t}_{c \to w} \\
        \mathbf{0}^\top & 1
    \end{bmatrix},
\end{equation}
where $\mathbf{R}_{c \to w} \in \mathbb{R}^{3 \times 3}$ is the rotation matrix and $\mathbf{t}_{c \to w} \in \mathbb{R}^3$ is the translation vector. The world-to-camera matrix $\mathbf{T}_{w \to c}$
used for 3D-to-2D projection is obtained by matrix inversion:
\begin{equation}
\mathbf{T}_{w \to c} = \mathbf{T}_{c \to w}^{-1}.
\end{equation}                           

\subsection{Geometric Feature Extraction}
Following the geometric alignment formulation in the Method section, we extract patch-level geometric features $f_g(I) \in \mathbb{R}^{T \times M \times S \times D_g}$ from multi-view images using the frozen geometric foundation model WorldMirror \cite{liu2025worldmirror} as $f_g$, where $T$ denotes the number of temporal frames, $M$ denotes the number of camera views, $S$ denotes the number of patch tokens per view, and $D_g=2048$ denotes the geometric feature dimension.                                           

\textbf{Camera Prior Extraction.} \ Given multi-view images $\mathbf{I} \in \mathbb{R}^{M \times 3 \times H \times W}$ along with their camera intrinsics   
$\mathbf{K} \in \mathbb{R}^{M \times 3 \times 3}$ and extrinsics $\mathbf{T} \in \mathbb{R}^{M \times 4 \times 4}$, the geometric foundation model first       
encodes these inputs into spatial-geometric priors $\mathcal{P}$ that encapsulate camera poses, depth estimates, and intrinsic parameters:

\begin{equation}
    \mathcal{P} = f_{\text{prior}}\left(\mathbf{I}, \mathbf{K}, \mathbf{T}\right).
\end{equation} 

\textbf{Geometric Feature Encoding.} \ The visual geometry transformer then processes the images conditioned on these priors. Following the model's conditional design, we utilize camera pose and intrinsic information (indicated by condition flags $\mathbf{c} = [1, 0, 1]$ for camera pose, depth, and intrinsics respectively) while excluding depth supervision:
\begin{equation}
    f_g(I) = \text{WorldMirror}\left(\mathbf{I}, \mathcal{P}; \mathbf{c}\right).
\end{equation}

Since $f_g$ remains frozen throughout training, we pre-compute and offline-cache $f_g(I)$ for all training samples, enabling direct loading during training without repeated inference costs. This design substantially reduces training time and GPU memory overhead. 

\section{Inference Latency}
We measured the inference latency per module of Latent-WAM on a single A100 GPU for a single batch. The inference latency of each module is shown in Table~\ref{tab:latency}. Latency was evaluated after three warm-up iterations, with the final number averaged over 10 forward passes.

\begin{table}[h]
    \centering
    \caption{\textbf{Per-module inference latency of Latent-WAM}}
    \label{tab:latency}
    \resizebox{0.45\textwidth}{!}{
        \begin{tabular}{c|cccc}
        \toprule
        \textbf{Module} \ & \ \textbf{Parameters} &  & \textbf{Latency} \ \\
        \midrule
        World Encoder & 86.6M &  & 100ms \\
        Trajectory Decoder \ & 8.4M &  & 6ms\\
        All modules & 104M &  & 107ms\\
        \bottomrule
        \end{tabular}
    }
\end{table}

\begin{figure*}
  \centering                         
  \includegraphics[width=0.85\textwidth]{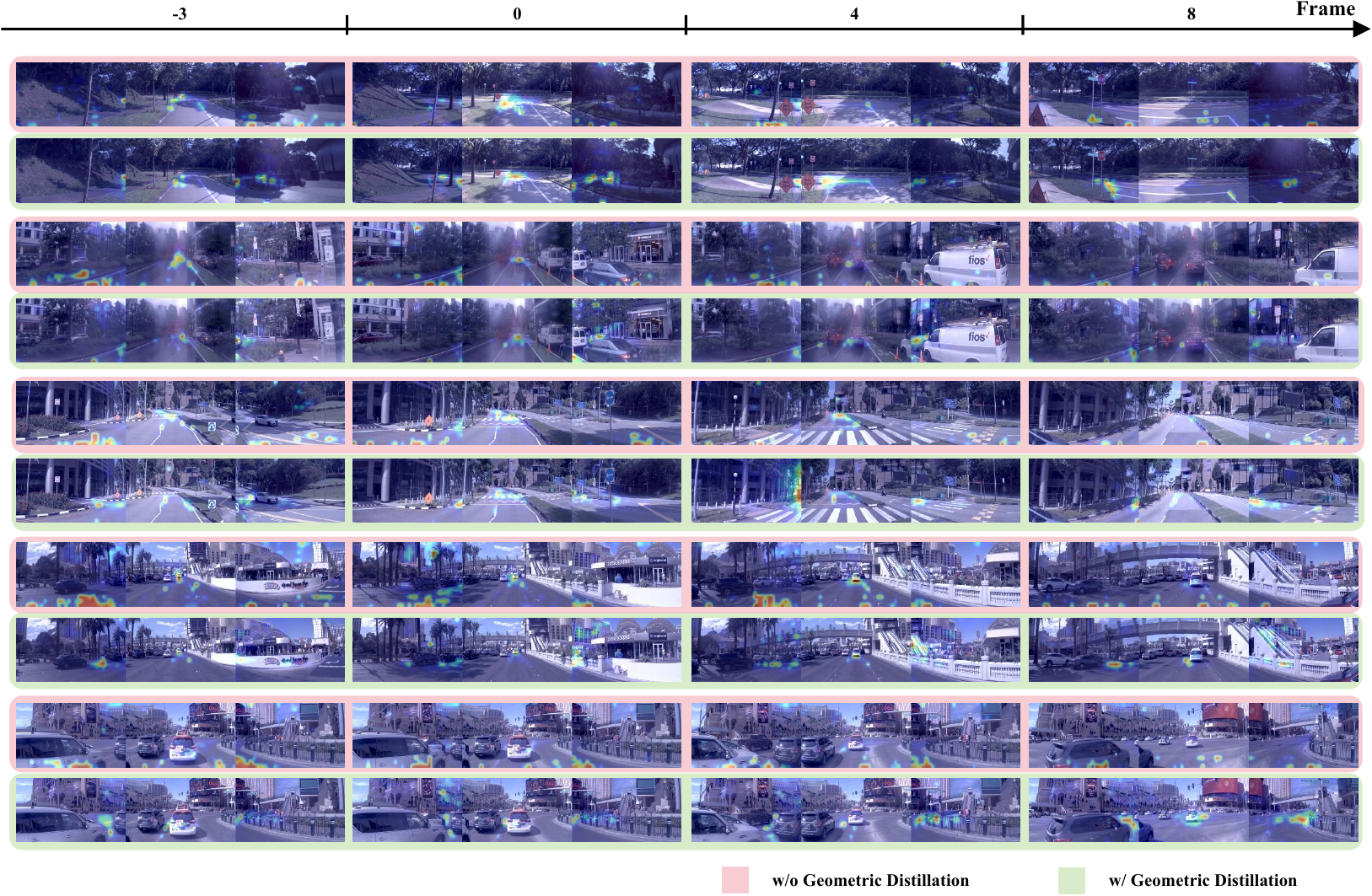}
  \caption{Visualization of attention maps between scene tokens and image patches across consecutive frames. All groups depict \textbf{going-forward scenarios}. Geometric distillation yields more focused attention on task-relevant regions.}
  \label{fig:attention_append_1}

  \includegraphics[width=0.85\textwidth]{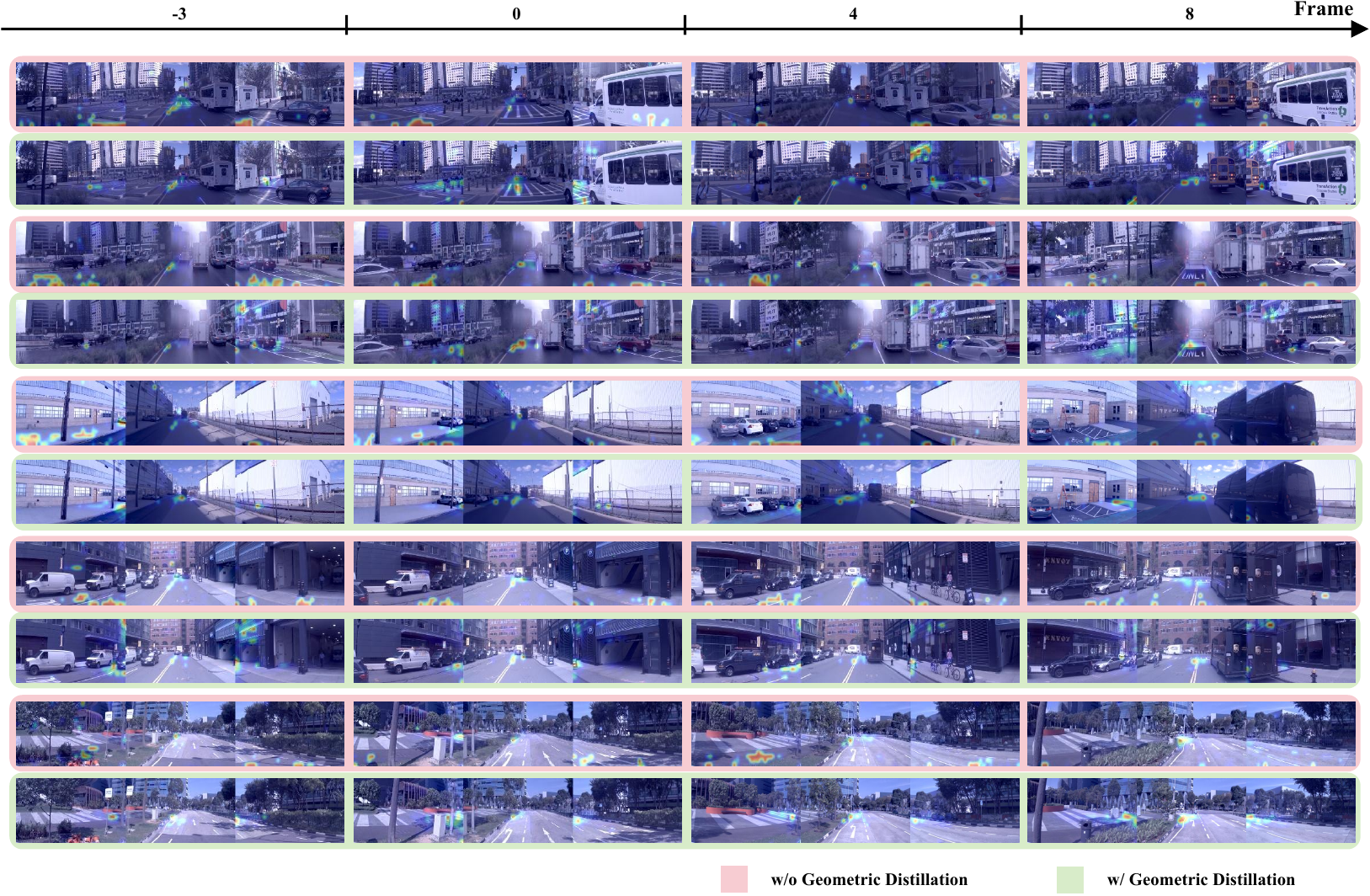}
  \caption{Visualization of attention maps between scene tokens and image patches across consecutive frames. All groups depict \textbf{lane-changing scenarios}. Geometric distillation yields more focused attention on task-relevant regions.}
  \label{fig:attention_append_2}
\end{figure*}

\section{More Visualization}
  We provide additional qualitative results to complement the quantitative analysis.

  \subsection{Trajectory Planning}
  
  Fig.~\ref{fig:traj_append_1} presents additional trajectory planning comparisons on NAVSIM across diverse driving scenarios. The green line denotes the human expert     
  trajectory and the yellow line represents the predicted trajectory. Compared to Epona~\cite{zhang2025epona} and World4Drive~\cite{zheng2025world4drive}, Latent-WAM consistently produces trajectories
  that more closely follow the expert demonstration while maintaining safer distances from surrounding vehicles, demonstrating the robustness of our approach across different traffic conditions.         
                                                            
  \subsection{Attention Map Visualization}
   
\begin{figure*}
    \centering
    \includegraphics[width=0.85\linewidth]{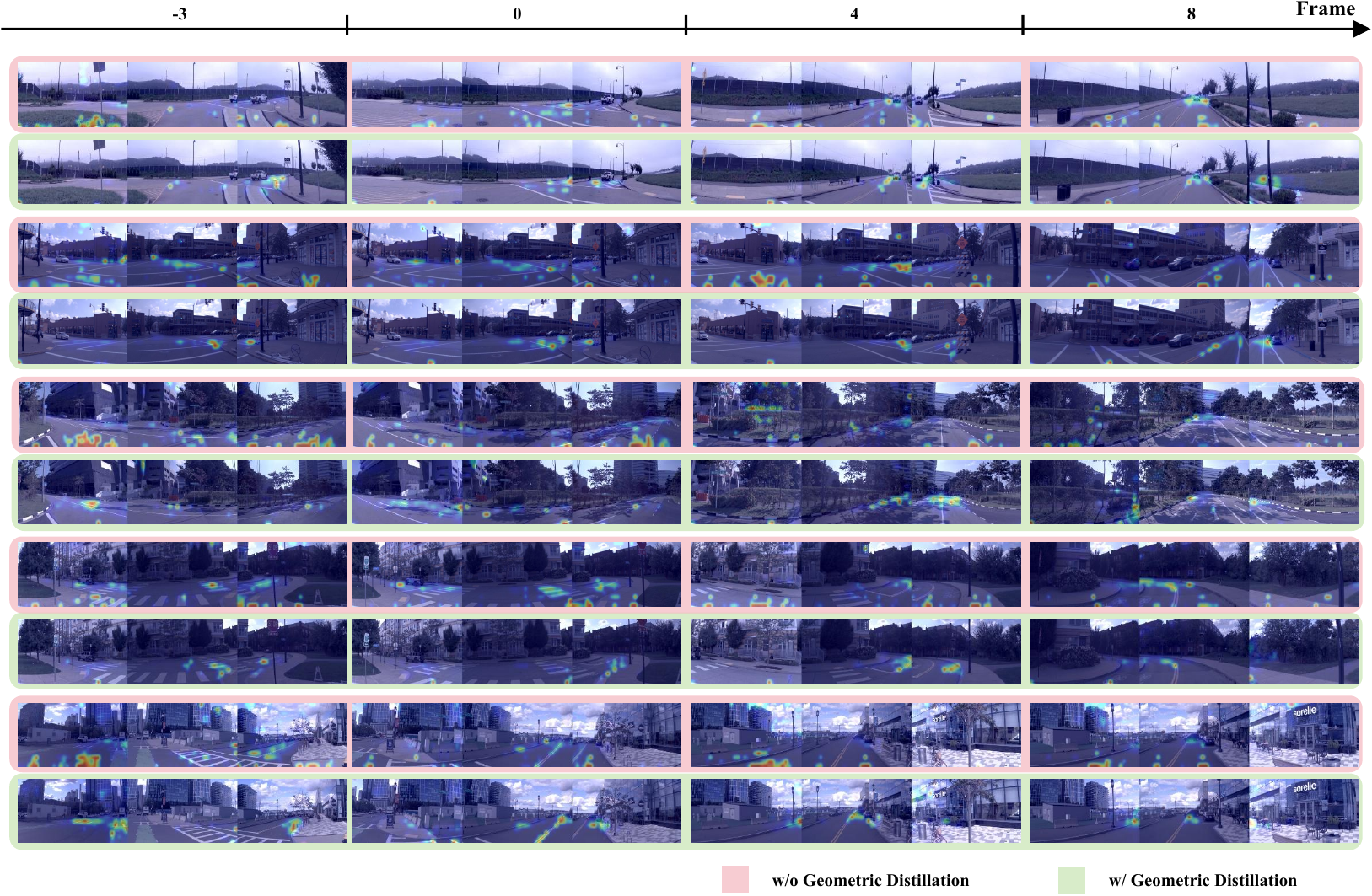}
    \captionof{figure}{Visualization of attention maps between scene tokens and image patches across consecutive frames. All groups depict \textbf{turn-right scenarios}. Geometric distillation yields more focused attention on task-relevant regions.}
    \label{fig:attention_append_3}
    
    \includegraphics[width=0.85\linewidth]{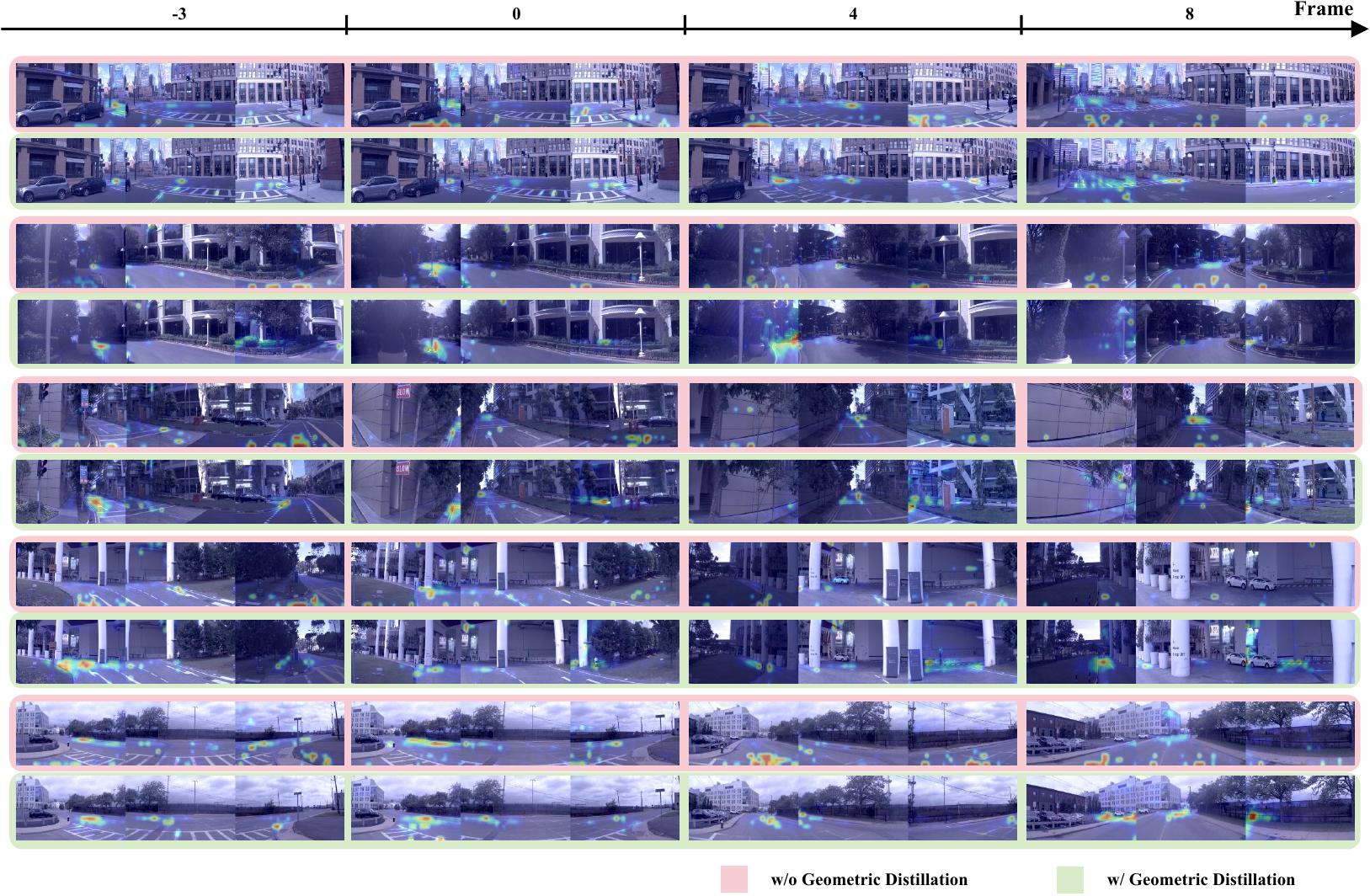}
    \captionof{figure}{Visualization of attention maps between scene tokens and image patches across consecutive frames. All groups depict \textbf{turn-left scenarios}. Geometric distillation yields more focused attention on task-relevant regions.}
    \label{fig:attention_append_4}
\end{figure*}

  Figs.~\ref{fig:attention_append_1}--\ref{fig:attention_append_4} present additional attention map visualizations between scene tokens and image patches across consecutive frames, covering four driving 
  scenarios: going-forward (Fig.~\ref{fig:attention_append_1}), lane-changing (Fig.~\ref{fig:attention_append_2}), right-turning (Fig.~\ref{fig:attention_append_3}), and left-turning
  (Fig.~\ref{fig:attention_append_4}). Consistent with the observations in the main paper, our model with geometric distillation yields more focused attention on task-relevant regions such as lane       
  markings and drivable areas, while the baseline without distillation produces scattered attention over irrelevant backgrounds. Furthermore, the attention distribution is strongly correlated with the
  driving intent across all scenarios, confirming that our SCWE learns to selectively focus on spatial information conditioned on the driving context.

  \subsection{HUGSIM Benchmark Visualization}
   
\begin{figure*}
    \centering
    \includegraphics[width=0.7\linewidth]{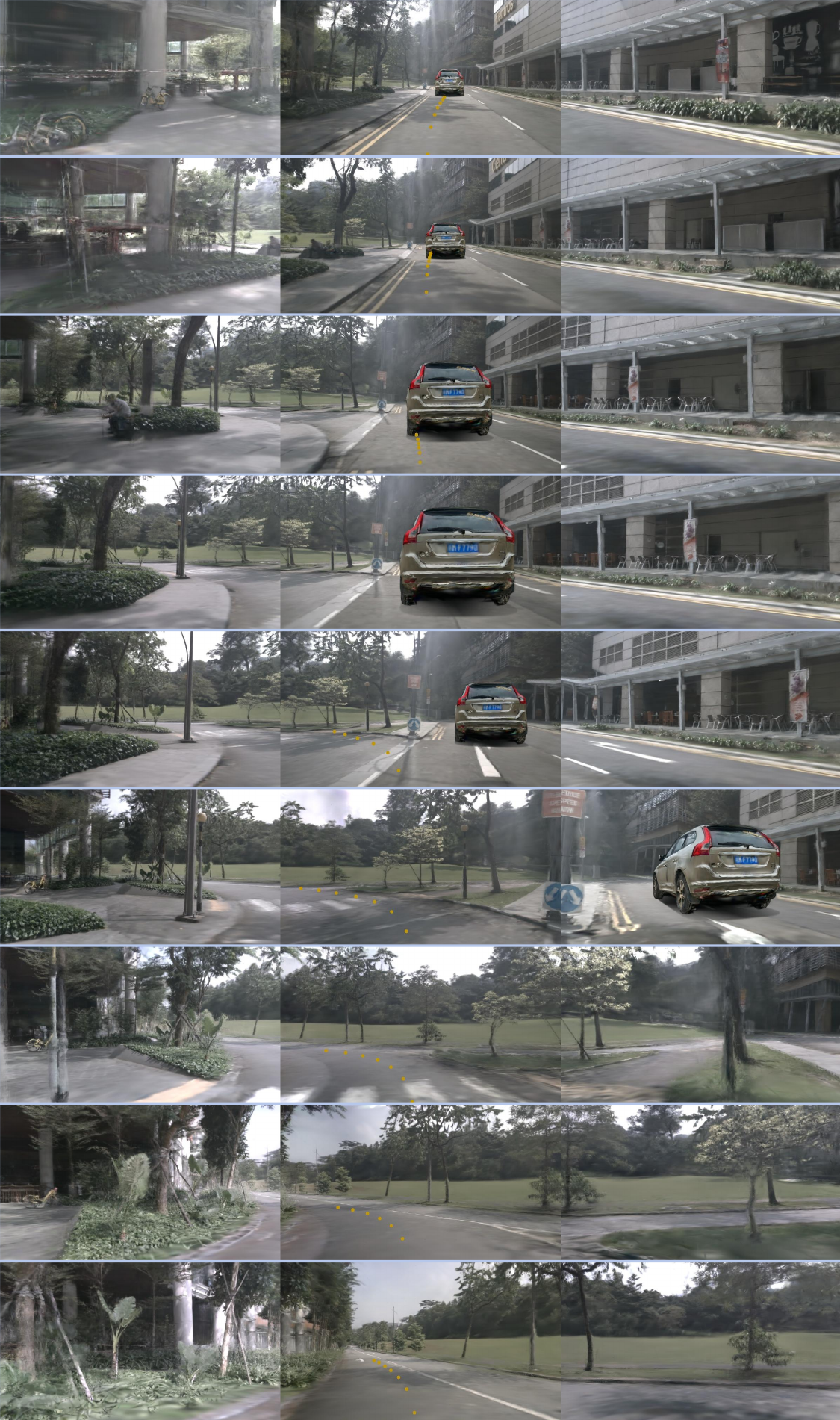}
    \captionof{figure}{Visualization of trajectory planning on HUGSIM benchmark(nuScenes \cite{nuscenes})}
    \label{fig:nuScenes}
\end{figure*}

\begin{figure*}
    \centering
    \includegraphics[width=0.7\linewidth]{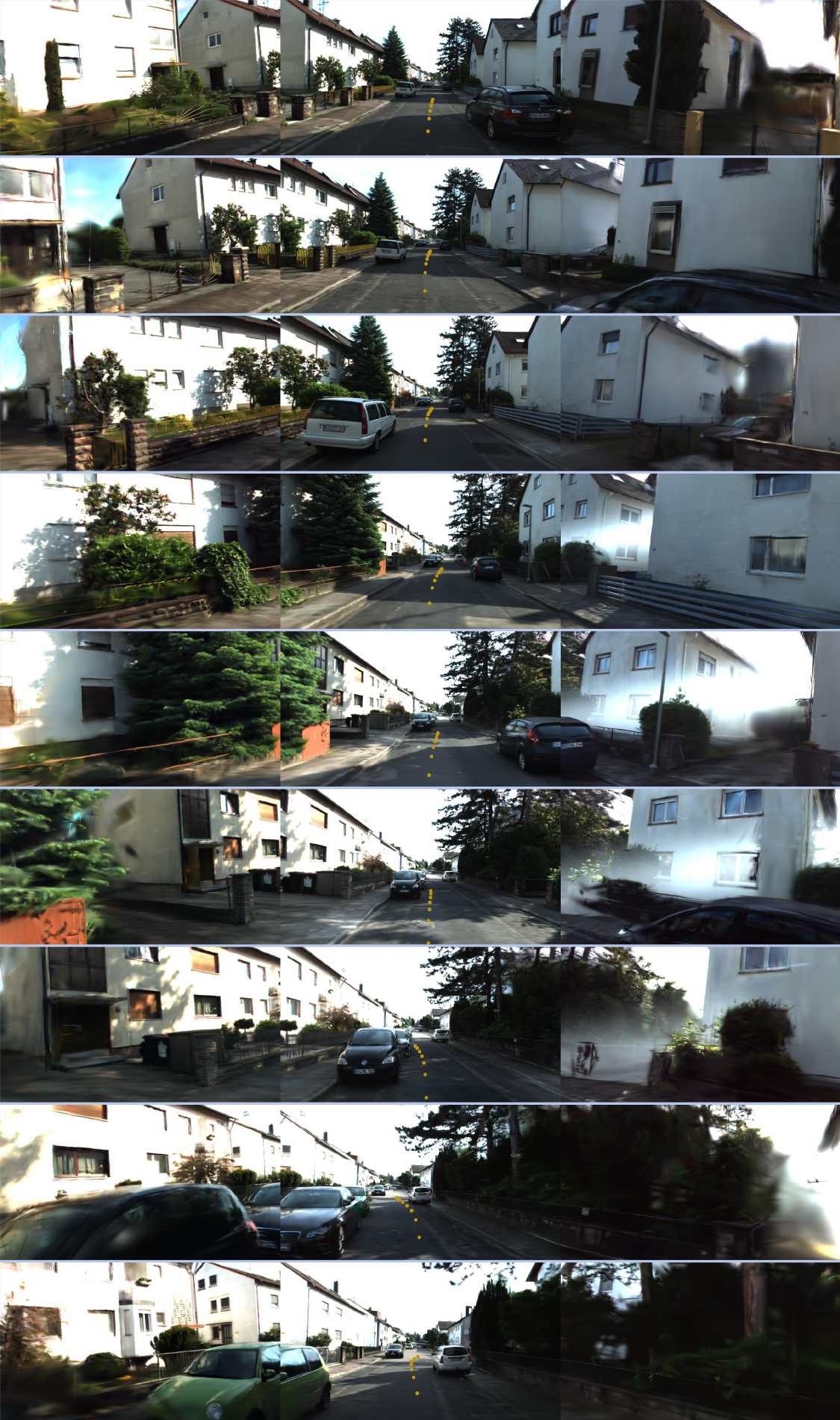}
    \captionof{figure}{Visualization of trajectory planning on HUGSIM benchmark(KITTI-360 \cite{kitti360})}
    \label{fig:kitti1}
\end{figure*}

\begin{figure*}
    \centering
    \includegraphics[width=0.7\linewidth]{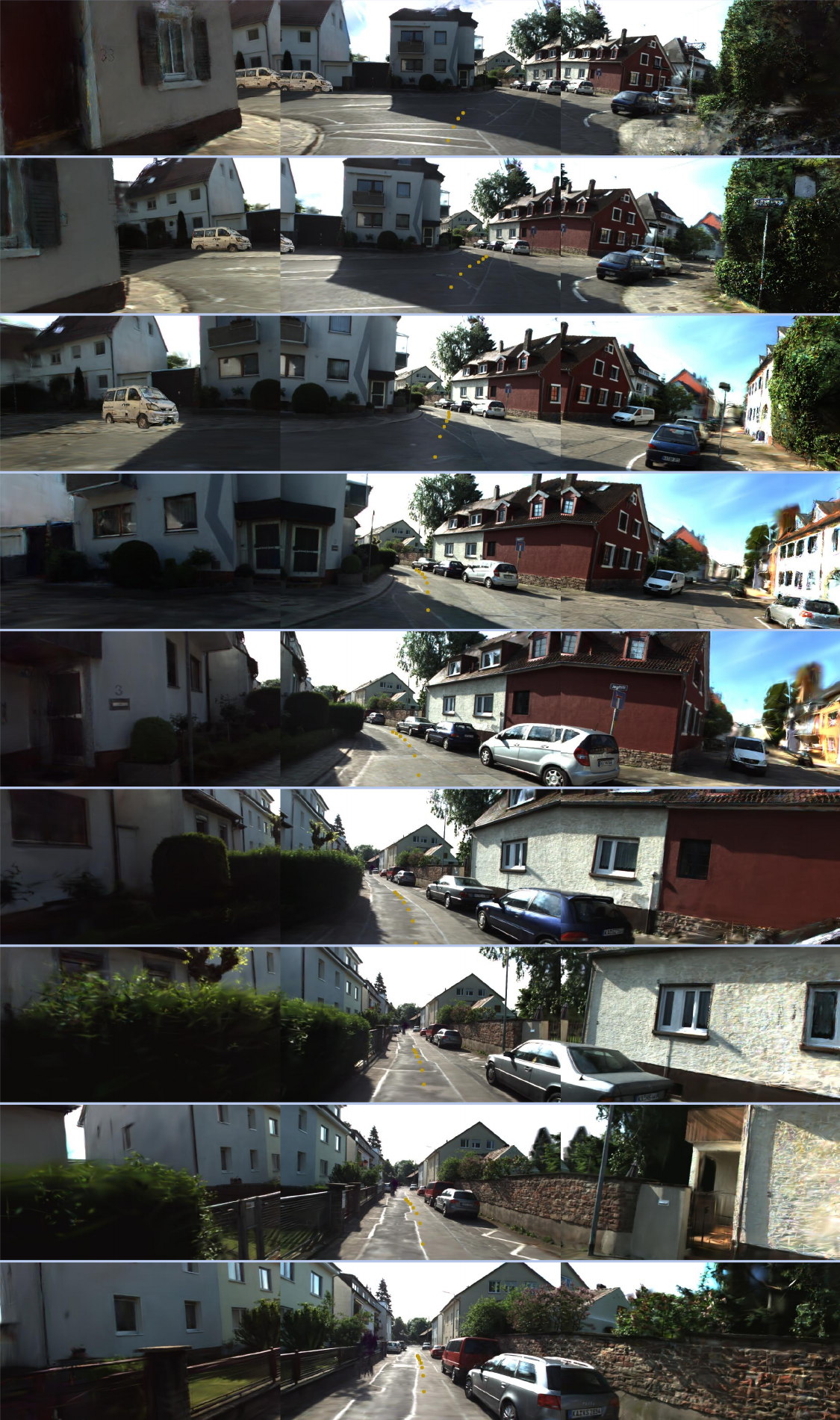}
    \captionof{figure}{Visualization of trajectory planning on HUGSIM benchmark(KITTI-360)}
    \label{fig:kitti2}
\end{figure*}

\begin{figure*}
    \centering
    \includegraphics[width=0.7\linewidth]{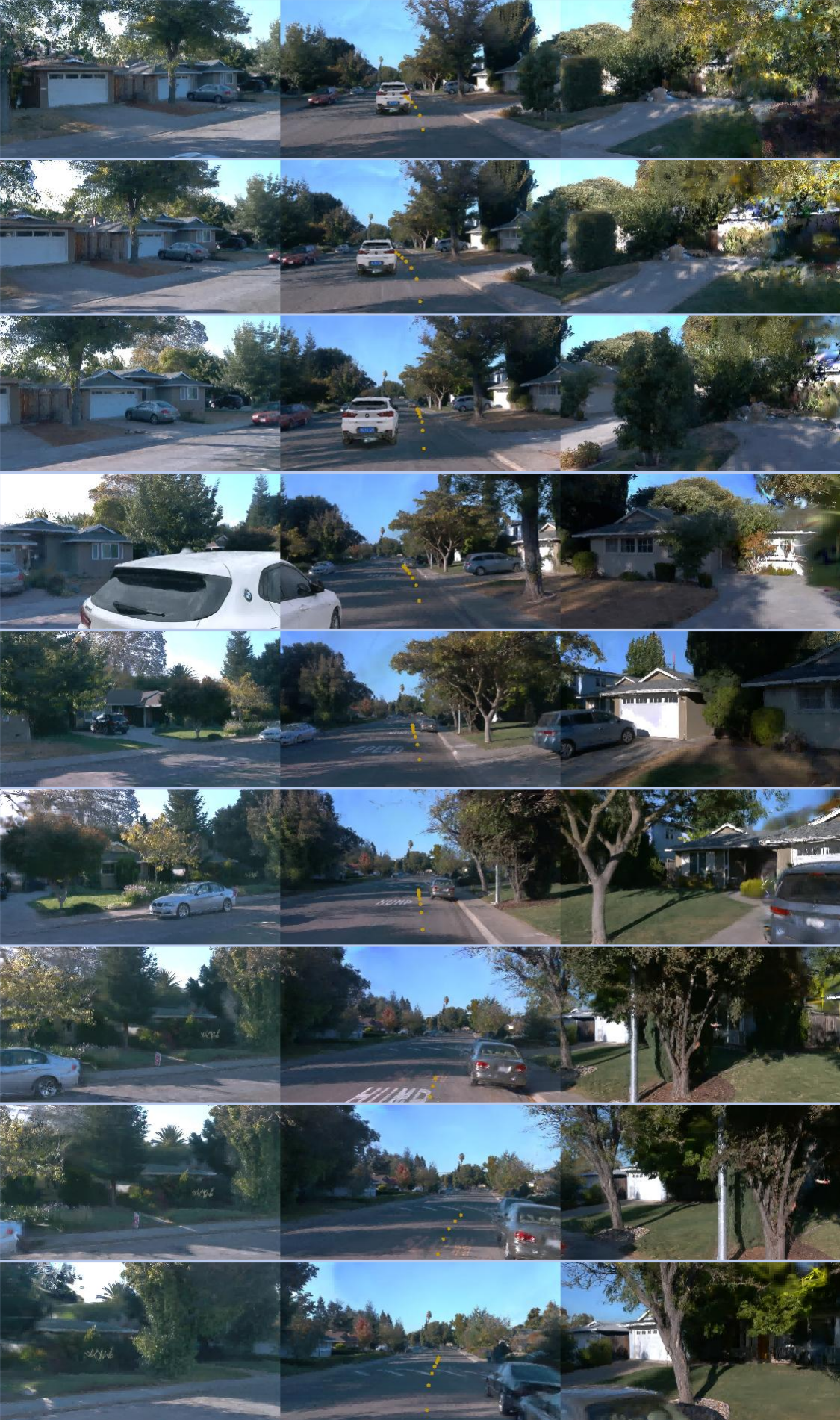}
    \captionof{figure}{Visualization of trajectory planning on HUGSIM benchmark(Waymo \cite{waymo})}
    \label{fig:waymo}
\end{figure*}

\begin{figure*}
    \centering
    \includegraphics[width=0.7\linewidth]{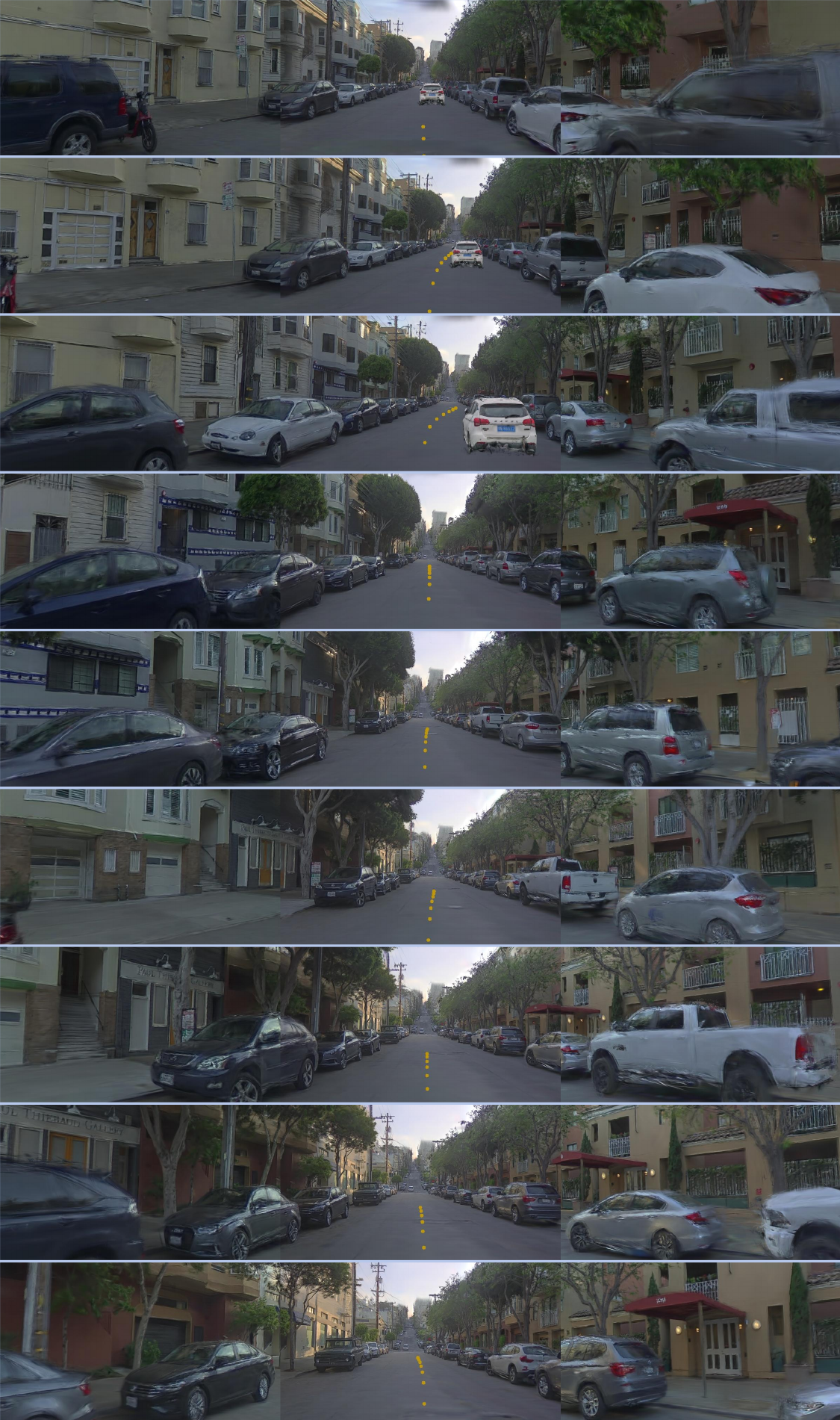}
    \captionof{figure}{Visualization of trajectory planning on HUGSIM benchmark(PandaSet \cite{pandaset})}
    \label{fig:pandaset}
\end{figure*}

  Figs.~\ref{fig:nuScenes}--\ref{fig:pandaset} present trajectory planning visualizations on the HUGSIM benchmark across four datasets: nuScenes~\cite{nuscenes} (Fig.~\ref{fig:nuScenes}),                
  KITTI-360~\cite{kitti360} (Figs.~\ref{fig:kitti1} and \ref{fig:kitti2}), Waymo~\cite{waymo} (Fig.~\ref{fig:waymo}), and PandaSet~\cite{pandaset} (Fig.~\ref{fig:pandaset}). Despite being trained
  exclusively on NAVSIM without any fine-tuning, Latent-WAM demonstrates strong zero-shot generalization capability, producing reasonable and safe trajectories across diverse driving environments        
  reconstructed from different real-world datasets.        

\end{document}